\documentclass[journal,onecolumn,11pt]{IEEEtran}

\usepackage{setspace}
\usepackage{blindtext}
\usepackage{geometry}
\usepackage[T1]{fontenc}
\usepackage[latin9]{inputenc}
\usepackage{amsthm}
\usepackage{amsmath}
\usepackage{amssymb}
\usepackage{graphicx}
\usepackage{balance}
\usepackage{epstopdf}
\usepackage{units}
\usepackage[nospace]{cite}
\usepackage{mathrsfs}
\usepackage{algorithm}
\usepackage{relsize}
\usepackage[scr=zapfc,scrscaled=1.2]{mathalfa}
\usepackage{xcolor}
\usepackage{tcolorbox}
\usepackage{float}

\theoremstyle{Lemma}

\def \T{\mathsf{T}}
\def\H{\mathsf{H}}
\newcommand*{\p}[1]{\mathsf{p} \left\{ #1 \right\}}
\newcommand*{\tr}[1]{\mathsf{tr} \left\{ #1 \right\}}
\newcommand*{\E}[1]{\mathsf{E}\left\{ #1 \right\}}

\DeclareMathAlphabet\mathbfcal{OMS}{cmsy}{b}{n}
\newcommand*{\diag}[1]{\mathsf{diag}\left\{#1\right\}}

\makeatletter

\theoremstyle{plain}
\newtheorem{thm}{\protect\theoremname}
\theoremstyle{remark}
\newtheorem{rem}[thm]{\protect\remarkname}

\makeatother

\providecommand{\remarkname}{Remark}
\providecommand{\theoremname}{Theorem}
\graphicspath{{Figures/}}

\definecolor{ICL}{rgb}{0.08, 0.3378 0.6711}

\newtcolorbox[auto counter]{bitbox}{colback=ICL!36!white,colframe=ICL!36!white,coltitle=black,float=h}

\begin{document}
	
\title{The $\mathbb{HR}$-Calculus: Enabling Information Processing with Quaternion Algebra}
	
\author{{\small{Danilo P. Mandic, Sayed Pouria Talebi, Clive Cheong Took,\\Yili Xia, Dongpo Xu, Min Xiang, and Pauline Bourigault}}}
	
\maketitle
	
\IEEEpeerreviewmaketitle
	
\vspace{-1.8cm}

\section*{Abstract}

This article presents a tutorial overview of the $\mathbb{HR}$-calculus. The article presents the basic formulation of gradient for quaternion-valued functions, useful derivative rules, and their application in formulation of adaptive information processing techniques. Moreover, advantages of using quaternion-valued processing techniques, as compared to their real and complex-valued counterparts, are demonstrated in a number applications ranging from power system monitoring to quantum computing.\footnote{This work was supported in~part by the Young Scientist Fund of National Natural Science Foundation of China (No.~62101017).}
 
\section{Introduction}

Despite their impressive versatility in modelling real-world phenomena, adaptive information processing techniques specifically designed for quaternion-valued signals have only recently come to the attention of the machine learning, signal processing, and control communities~\cite{Cyrus,QuaternionControl,MeJ,QuantComplier,QML1,QML2}. The most important development in this direction is introduction of the $\mathbb{HR}$-calculus that provides the required mathematical foundation for deriving adaptive information processing techniques directly in the quaternion domain. This article, provides a tutorial overview on derivation of key concepts in $\mathbb{HR}$-calculus that enable quaternion-valued adaptive information processing techniques. Importantly,  the use of these techniques in a number of real-world applications is demonstrated. This article presents a general overview with detailed mathematical derivations and complementary application examples 
provided in the Supplementary Material (SM). However, before furthering the discussion on quaternions, their associate division algebra, and the $\mathbb{HR}$-calculus, it is important to introduce the nomenclature.

\noindent\textbf{\textit{Nomenclature}}: Scalars, column vectors, and matrices are denoted respectively by lowercase, bold lowercase, and bold uppercase letters. The reminder of nomenclature is summarised as follows:
\begin{IEEEdescription}
\item[$\mathbb{R}$, $\mathbb{R}^{+}$]\hspace{1.2cm}real and positive real numbers
\item[$\mathbb{C}$, $\mathbb{H}$, $\mathbb{N}$]\hspace{1.2cm}complex, quaternion, and natural numbers
\item[$\Re\{\cdot\}$, $\Im\{\cdot\}$]\hspace{1.2cm}operators returning the real and imaginary components
\item[$\{\imath,\jmath,\kappa\}$]\hspace{1.2cm}imaginary units spanning the imaginary subspace of $\mathbb{H}$
\item[$\Im_{\chi}\left(\cdot\right)$]\hspace{1.2cm}operator returning the imaginary component alongside $\chi\in\{\imath,\jmath,\kappa\}$
\item[$\|\cdot\|$]\hspace{1.2cm}second-order norm
\item[$\mathbf{I}$]\hspace{1.2cm}identity matrix of appropriate size
\item[$\wedge$]\hspace{1.26cm}logical conjunction 
\item[$(\cdot)^{*}$]\hspace{1.2cm}conjugate operator. 
\item[$\left(\cdot\right)^{\T}$, $\left(\cdot\right)^{\H}$]\hspace{1.2cm}transpose, and transpose Hermitian operators
\item[$e$]\hspace{1.2cm}Euler's number
\item[$\E{\cdot}$] \hspace{1.2cm}statistical expectation operator
\item[$\partial$, $\mathsf{d}$]\hspace{1.2cm}partial and total differential operators
\item[$\nabla_{\chi}$]\hspace{1.2cm}gradient operator with respect to $\chi$
\item[$\diag{\cdot}$]\hspace{1.2cm}constructs a block-diagonal matrix from its entries  
\item[$\p{\cdot}$, $\tr{\cdot}$]\hspace{1.2cm}spectral radius  and trace operators
\end{IEEEdescription}

\subsection{Overview of Quaternions}

Concerned with representing rotations in three-dimensional space, in a manner analogous to that achieved via complex-valued numbers for rotations in two-dimensional spaces, Sir William Rowan Hamilton formulated the basis of quaternion algebra. A quaternion number $q\in\mathbb{H}$, consists of a real part, $\Re\{q\}$, and a three-dimensional imaginary part, $\Im\{q\}$, which comprises three components, $\Im_{\imath}\{q\}$, $\Im_{\jmath}\{q\}$, and $\Im_{\kappa}\{q\}$. Thus, $q\in\mathbb{H}$ can be expressed as~\cite{QuatBook}
\begin{equation}
q =\Re\{q\}+ \Im\{q\}=\Re\{q\} + \Im_{\imath}\{q\} + \Im_{\jmath}\{q\} + \Im_{\kappa}\{q\}=q_{r} + \imath q_{\imath} + \jmath q_{\jmath} + \kappa q_{k}
\label{eq:QuatStructure}
\end{equation}
where $\{q_{r},q_{\imath},q_{\jmath},q_{\kappa}\}\subset\mathbb{R}$, with $\{\imath,\jmath,\kappa\}$ admitting the product rules~\cite{QuatBook}
\begin{equation}
\imath\jmath=\kappa,\hspace{0.12cm}\jmath\kappa=\imath,\hspace{0.12cm}\kappa\imath=\jmath,\hspace{0.12cm}\imath^{2}=\jmath^{2}=\kappa^{2}=\imath\jmath\kappa=-1.
\label{eq:ProductRule}
\end{equation} 
The product of $q_{1}, q_{2}\in\mathbb{H}$ can be found through component-wise multiplication of their real-valued constituents using the product rules in \eqref{eq:ProductRule}. Although associative, as a direct consequence of \eqref{eq:ProductRule}, product operations are not commutative in the quaternion domain, unless under strict conditions, such as $\Im\{q_{1}\}=0$ or $\Im\{q_{2}\}=0$. For example, note that $\imath\jmath=\kappa$ while $\jmath\imath=\kappa\imath\imath=-k$, where we have used $\jmath=\kappa\imath$ form \eqref{eq:ProductRule}. 

\begin{rem}
Selection of the units that span the imaginary subspace of $\mathbb{H}$ is not unique. Indeed, a convenient change of these units can be used to simplify derivation and analysis of quaternion-valued adaptive processing techniques, see~\cite{PouriaPhD,Phase3Freq,Mej}. For example, change of basis is used in power system modelling and derivation of the multiplication rule, which are respectively presented in Section~\ref{Sec:Kalman} and Section~\ref{Sec:MRule}. 
\end{rem}

The quaternion conjugate is defined as $q^{*}=\Re(q)-\Im(q)$, which allows for an inverse to be defined as
\begin{equation}
\forall q\in\mathbb{H}\setminus\{0\}: q^{-1} = \frac{q^{*}}{\left\|q\right\|^{2}}
\end{equation}
with $\left\|q\right\|$ denoting the second-order norm of $q$, that is, $\left\|q\right\|=\sqrt{qq^{*}}=\sqrt{ q^{2}_{r} + q^{2}_{\imath} + q^{2}_{\jmath} + q^{2}_{\kappa}}$. Form the definition of the quaternion conjugate it follows that
\begin{equation}
\forall q_{1},q_{2}\in\mathbb{H}:\hspace{0.12cm}\left(q_{1}q_{2}\right)^{*}=q^{*}_{2}q^{*}_{1}.
\label{eq:RevMulConj}
\end{equation}

Now, consider $q\in\mathbb{H}\backslash\{0\}$, from \eqref{eq:ProductRule} and \eqref{eq:QuatStructure} it follows that $\Im^{2}\{q\}\in\left(-\infty,0\right)$. Therefore, $\xi=\Im\{q\}/\|\Im\{q\}\|$ represents a unit imaginary, i.e. $\xi^{2}=-1$. The basis $\left\{1,\xi\right\}$ span a complex-valued subspace in $\mathbb{H}$ that contains $q$. In this subspace, 
$q$ can be formulated in the polar representation~\cite{QuatBook,PouriaPhD}
\begin{equation}
q =\|q\|e^{\xi \theta}=\|q\|\big(\cos\left(\theta\right)+\xi \sin\left(\theta\right)\big)\hspace{0.26cm}\text{where}\hspace{0.26cm}\text{and}\hspace{0.26cm}\theta=\text{atan}\left(\frac{\|\Im(q)\|}{\Re(q)}\right).
\end{equation}
Thus, it follows that $\sin(\cdot)$ and $\cos(\cdot)$ functions can be expressed as
\begin{equation}
\sin\left(\theta\right)=\frac{1}{2\xi}\left(e^{\xi \theta} - e^{- \xi \theta}\right)\hspace{0.5cm}\text{and}\hspace{0.5cm} \hspace{0.5em}\cos\left(\theta\right)=\frac{1}{2}\left(e^{\xi \theta} + e^{- \xi \theta}\right).
\label{eq:QuaternionSinCos}
\end{equation}
This formulation allows for an elegant representation for three-dimensional rotations, which is explored in the sequel.

\subsection{Three-Dimensional Rotations}

\label{Sec:QuaternionRotation}

\begin{figure}[h!]
\centering
\includegraphics[width=0.9\linewidth,trim=0cm 0.6cm 0cm 0cm]{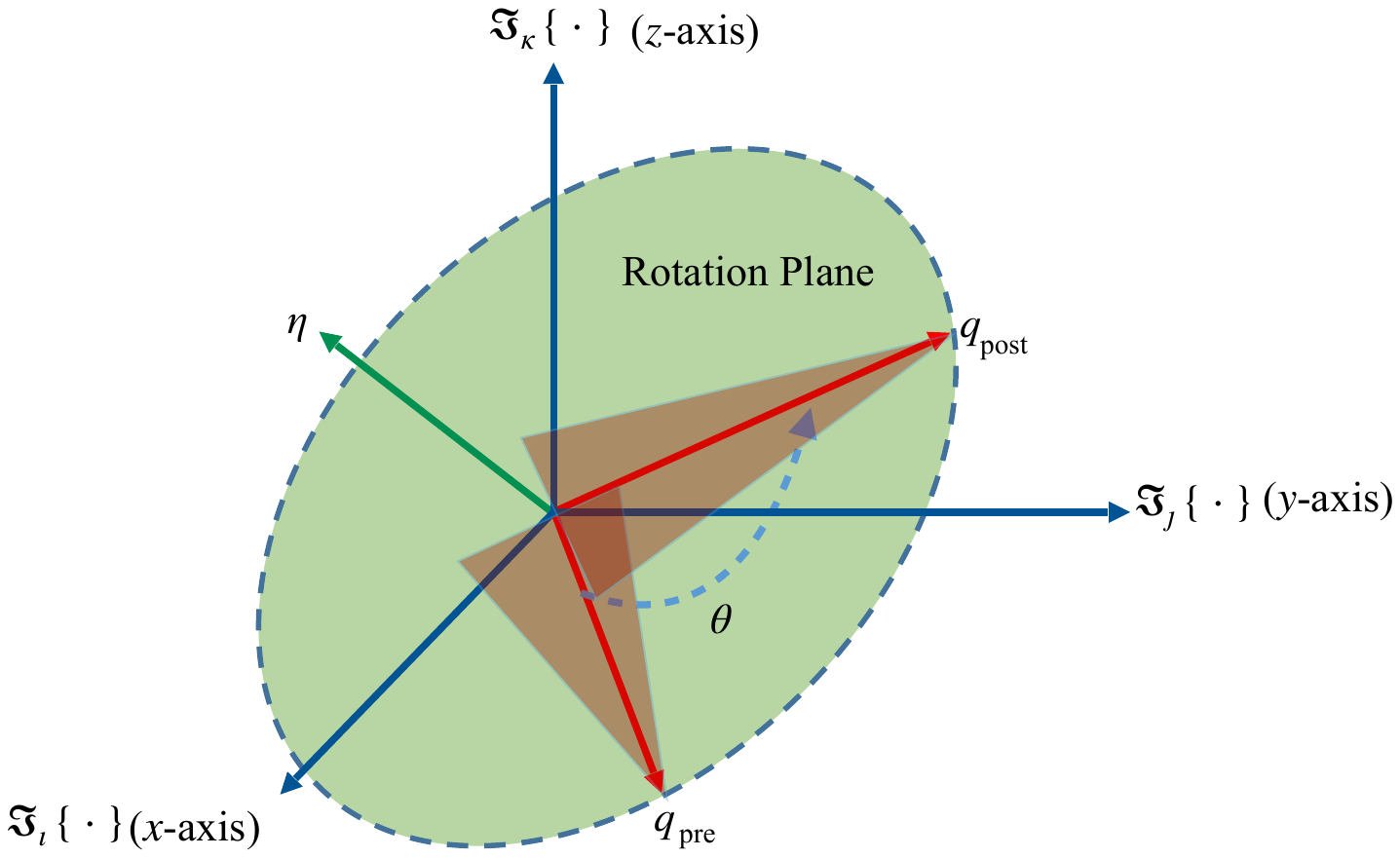}
\caption{Schematic of a rotation around $\eta$ by an angle of $\theta$, with $q_{\text{pre}}$ and $q_{\text{post}}$ pointing to the pre- and post-rotation orientation of the object in question.}
\label{Fig:Rotation}
\end{figure}

Consider a single right-hand rotation of an object by an angle of $\theta$ degrees about an axis $\eta$, as shown in Figure~\ref{Fig:Rotation}. The three-dimensional Cartesian coordinate system can be considered as representing the imaginary sub-space of quaternions, so that $\eta=\left(\eta_{x},\eta_{y},\eta_{z}\right)$ can be presented as $\eta=\imath\eta_{x}+\jmath\eta_{y}+\kappa\eta_{z}$. In this setting, the vector presenting 
the pre- and post-rotation orientation of the object, $q_{\text{pre}}$ and $q_{\text{post}}$, are related as~\cite{Kuipers}
\begin{equation}
q_{\text{post}}=\mu q_{\text{pre}}\mu^{-1}\hspace{0.36cm}\text{with}\hspace{0.36cm}\mu=e^{\eta\frac{\theta}{2}}=\cos\left(\frac{\theta}{2}\right)+\eta\sin\left(\frac{\theta}{2}\right).
\label{eq:QuaternionRotation}
\end{equation}

\subsection{Main Advantages of Quaternions}

Up to this point, focus of the information processing community has mainly been on the use of real-valued vector algebras. In principle, these techniques can handle data of any dimensions. Thus, the validity of deriving information processing techniques in the quaternion domain is sometimes questioned. The answer to these questions can be classified as follows: 

\begin{itemize}
	
\item\textbf{Dealing with Rotations}: Representing rotations via quaternions, as compared to Euler angles and real-valued rotation matrix algebras, holds a number of advantages. First, real-valued rotation matrices lose a degree of freedom as one of the rotation angles reaches $\frac{\pi}{2}$~\cite{PouriaPhD}, whereas this is not the case for quaternions.\footnote{This is a phenomena referred to as algebraic gimbal-lock~\cite{PouriaPhD,Kuipers}.} Second, real-valued rotation matrices must have a unit determinant. This is hard to guarantee after many rotations have been computed due to the finite accuracy of computer operations. This problem does not arise in the quaternion representation~\cite{Cyrus,PouriaPhD}. This vantage point is used in aerospace~\cite{Kuipers,PouriaPhD}, body motion tracking~\cite{QRKHS}, and modelling of quantum computing gates~\cite{QuantComplier}. 

\item\textbf{Solving Differential Equations}: In engineering, behaviour of most physical phenomena is explainable via a set of real-valued differential equations. Finding, even numerical solutions to these equations in the real domain is not a straightforward affair. However, using transformations such as the Fourier and Z-transforms allow transition to a new complex-valued domain, where solutions are attainable with relative ease. The quaternion domain accommodates for similar approaches in cases where the underlying signal is two or three-dimensional, e.g. multi-channel sound processing~\cite{3DS,MUSIC}, colour image processing~\cite{Image-2}, and communication techniques that adopt space-time-polarisation multiplexing~\cite{Comm1,Comm2}.

\end{itemize}   
Moreover, in a number of applications, such as power system modelling,  both of the raised issues come into play~\cite{Phase3Freq,3DElec,Rus,NotQ}. In these applications a straightforward notion of phase makes extracting and dealing with angles and frequency more efficient and their computations more accurate, e.g. see Section~\ref{Sec:Kalman} and SM-8. Finally, it should be noted that there seems to be an inherent relation between the formulation in \eqref{eq:QuaternionRotation} and the algebraic Lyapunov and Riccati recursions, two of the most fundamental equations in adaptive signal processing and control~\cite{Stengel}, which is further explored in SM-1.

\subsection{Generalising Rotations: Quaternion Involution and the Augmented Representation}
	
The operation in \eqref{eq:QuaternionRotation} was defined with a pure imaginary input/output in mind. However, a slight modification accommodates for a powerful tool in quaternion algebra. The involution\footnote{The strict definition of an involution in mathematical parlance is a mapping that is its own inverse. However, a richer construct has been shown to be useful in the context of quaternions~\cite{Q-Invo}.} of $q\in\mathbb{H}$ around $\zeta\in\mathbb{H}\setminus\{0\}$ is defined as~\cite{Q-Invo}
\begin{equation}
q^{\zeta}=\zeta q \zeta^{-1}=\zeta\Re\{q\}\zeta^{-1}+\zeta\Im\{q\} \zeta^{-1}=\Re\{q\}+\zeta\Im\{q\} \zeta^{-1}
\label{eq:Involution}
\end{equation}
where, $\Im\{q\}$ is rotated around $\zeta$ by $2\text{atan}\left(\frac{\|\Im\{\zeta\}\|}{\Re\{\zeta\}}\right)$ while $\Re\{q\}$ is unchanged. This somewhat resembles operation of the complex conjugate operator. Indeed, in a similar use case as for the complex conjugate operator, real-valued components of quaternions can be expressed using the linear and invertible mapping\footnote{The quaternion conjugate can also be formulated using involutions as $q^{*}=\frac{1}{2}\left(q^{\imath} + q^{\jmath} + q^{\kappa} - q \right)$.}~\cite{AQS,HR-Gradient}
\begin{equation}
\underbrace{\begin{bmatrix}\mathbf{q}\\\mathbf{q}^{\imath}\\\mathbf{q}^{\jmath}\\\mathbf{q}^{\kappa}\end{bmatrix}}_{\mathbf{q}^{a}}=\underbrace{\begin{bmatrix}\mathbf{I}&\imath\mathbf{I}&\jmath\mathbf{I}&\kappa\mathbf{I}\\\mathbf{I}&\imath\mathbf{I}&-\jmath\mathbf{I}&-\kappa\mathbf{I}\\\mathbf{I}&-\imath\mathbf{I}&\jmath\mathbf{I}&-\kappa\mathbf{I}\\\mathbf{I}&-\imath\mathbf{I}&-\jmath\mathbf{I}&\kappa\mathbf{I}\end{bmatrix}}_{\mathbf{A}}\begin{bmatrix}\mathbf{q}_{r}\\\mathbf{q}_{\imath}\\\mathbf{q}_{\jmath}\\\mathbf{q}_{\kappa}\end{bmatrix}
\label{eq:MappingHR}
\end{equation}
where $\mathbf{q}^{i}$, $\mathbf{q}^{j}$, and $\mathbf{q}^{k}$, denote the element-wise implementation of the involution, $\mathbf{A}^{{}^{-1}}=\frac{1}{4}\mathbf{A}^{\H}$, and $\mathbf{q}^{a}$ is referred to as the augmented quaternion vector. The $\mathbb{HR}$-calculus uses the relation in \eqref{eq:Involution} to establish a notion of gradient for quaternion-valued functions, in a manner similar to that achieved via the  $\mathbb{CR}$-calculus~\cite{CR,TheCR} for complex-valued functions. This approach allows most widely used information processing tools to be formulated in the quaternion domain. 

\section{The $\mathbb{HR}$-calculus}

The $\mathbb{HR}$-calculus~\cite{Cyrus,GenHR,HR-Gradient} presents an elegant solution to the problem of formularising the steepest descent direction of quaternion-valued functions. To this end, consider the quaternion-valued function $\mathrm{f}\left(\mathbf{q}\right):\mathbb{H}^{M}\rightarrow\mathbb{H}$ presented in an augmented format as 
\begin{equation}
\mathrm{f}^{a}(\mathbf{q}^{a})=\begin{bmatrix}\mathrm{f}(\mathbf{q}^{a}),\mathrm{f}^{\imath}(\mathbf{q}^{a}),\mathrm{f}^{\jmath}(\mathbf{q}^{a}),\mathrm{f}^{\kappa}(\mathbf{q}^{a})\end{bmatrix}^{\T}:\mathbb{H}^{4M}\rightarrow\mathbb{H}^{4}.
\label{eq:AugmentedFunction}
\end{equation}
On one hand, the total derivative of $\mathrm{f}\left(\mathbf{q}\right)=\mathrm{f}\left(\mathbf{q}_{r},\mathbf{q}_{\imath},\mathbf{q}_{\jmath},\mathbf{q}_{\kappa}\right)$ is given by 
\begin{equation}
\mathsf{d}\mathrm{f}=\mathsf{d}\mathbf{q}_{r}\frac{\partial\mathrm{f}}{\partial\mathbf{q}_{r}}+\mathsf{d}\mathbf{q}_{\imath}\frac{\partial\mathrm{f}}{\partial\mathbf{q}_{\imath}}+\mathsf{d}\mathbf{q}_{\jmath}\frac{\partial\mathrm{f}}{\partial\mathbf{q}_{\jmath}}+\mathsf{d}\mathbf{q}_{\kappa}\frac{\partial\mathrm{f}}{\partial\mathbf{q}_{\kappa}}
\label{eq:total-Q-diff-Real}
\end{equation}
while on the other hand, treating $\{\mathbf{q},\mathbf{q}^{\imath},\mathbf{q}^{\jmath},\mathbf{q}^{\kappa}\}$ as a set of ``\textit{algebraically}'' independent variables\footnote{Note that this is analogous to the $\mathbb{CR}$-calculus, where $z$ and $z^{*}$ are considered algebraically independent.} and considering the  augmented formulation in \eqref{eq:AugmentedFunction} we arrive at~\cite{Cyrus}
\begin{equation}
\mathsf{d}\mathrm{f}=\mathsf{d}\mathbf{q}\frac{\partial\mathrm{f}}{\partial\mathbf{q}} +\mathsf{d}\mathbf{q}^{\imath}\frac{\partial\mathrm{f}}{\partial\mathbf{q}^{\imath}}+\mathsf{d}\mathbf{q}^{\jmath}\frac{\partial\mathrm{f}}{\partial\mathbf{q}^{\jmath}}+\mathsf{d}\mathbf{q}^{\kappa}\frac{\partial\mathrm{f}}{\partial\mathbf{q}^{\kappa}}\cdot
\label{eq:total-Q-diff-Quaternion}
\end{equation}
Now, equating \eqref{eq:total-Q-diff-Real} and \eqref{eq:total-Q-diff-Quaternion} gives the so-called $\mathbb{HR}$-derivatives
\begin{equation}
\begin{bmatrix}
\frac{\partial\mathrm{f}(\mathbf{q},\mathbf{q}^{\imath},\mathbf{q}^{\jmath},\mathbf{q}^{\kappa})}{\partial\mathbf{q}}\\
\frac{\partial\mathrm{f}(\mathbf{q},\mathbf{q}^{\imath},\mathbf{q}^{\jmath},\mathbf{q}^{\kappa})}{\partial\mathbf{q}^{\imath}}\\
\frac{\partial\mathrm{f}(\mathbf{q},\mathbf{q}^{\imath},\mathbf{q}^{\jmath},\mathbf{q}^{\kappa})}{\partial \mathbf{q}^{\jmath}}\\
\frac{\partial\mathrm{f}(\mathbf{q},\mathbf{q}^{\imath},\mathbf{q}^{\jmath},\mathbf{q}^{\kappa})}{\partial \mathbf{q}^{\kappa}}
\end{bmatrix}
=\frac{1}{4}\mathbf{A}^{*}
\begin{bmatrix}
\frac{\partial\mathrm{f}(\mathbf{q}_{r},\mathbf{q}_{\imath},\mathbf{q}_{\jmath},\mathbf{q}_{\kappa})}{\partial \mathbf{q}_{r}}\\
\frac{\partial\mathrm{f}(\mathbf{q}_{r},\mathbf{q}_{\imath},\mathbf{q}_{\jmath},\mathbf{q}_{\kappa})}{\partial \mathbf{q}_{\imath}}\\
\frac{\partial\mathrm{f}(\mathbf{q}_{r},\mathbf{q}_{\imath},\mathbf{q}_{\jmath},\mathbf{q}_{\kappa})}{\partial \mathbf{q}_{\jmath}}\\
\frac{\partial\mathrm{f}(\mathbf{q}_{r},\mathbf{q}_{\imath},\mathbf{q}_{\jmath},\mathbf{q}_{\kappa})}{\partial \mathbf{q}_{\kappa}}
\end{bmatrix}
\label{eq:HR-derivatives}
\end{equation}
where \eqref{eq:MappingHR} was used to replace $\{\mathsf{d}\mathbf{q}_{r},\mathsf{d}\mathbf{q}_{\imath},\mathsf{d}\mathbf{q}_{\jmath},\mathsf{d}\mathbf{q}_{\kappa}\}$ with $\{\mathsf{d}\mathbf{q},\mathsf{d}\mathbf{q}^{\imath},\mathsf{d}\mathbf{q}^{\jmath},\mathsf{d}\mathbf{q}^{\kappa}\}$. The expression in \eqref{eq:HR-derivatives} establishes a direct relation between derivatives of $\mathrm{f}\left(\mathbf{q}_{r},\mathbf{q}_{\imath},\mathbf{q}_{\jmath},\mathbf{q}_{\kappa}\right)$ and derivates $\mathrm{f}\left(\mathbf{q}^{a}\right)$. On this basis a calculus for formulating derivatives of quaternion-valued function directly in the quaternion domain is established.  Particularly interesting in information processing applications, is  the conjugate derivative given by
\begin{equation}
\frac{\partial\mathrm{f}}{\partial \mathbf{q}^{*}}=\frac{1}{4}\left(\frac{\partial\mathrm{f}}{\partial\mathbf{q}_{r}}+\imath\frac{\partial\mathrm{f}}{\partial\mathbf{q}_{\imath}}+\jmath\frac{\partial\mathrm{f}}{\partial\mathbf{q}_{\jmath}}+\kappa\frac{\partial \mathrm{f}}{\partial\mathbf{q}_{\kappa}}\right)
\label{eq:DiffQuaternion}
\end{equation}
which indicates the direction of maximum rate of change in $\mathrm{f}(\cdot)$. Thus, presenting the gradient operator as
\begin{equation}
\begin{bmatrix}
\frac{\partial \mathrm{f} (\mathbf{q},\mathbf{q}^{i},\mathbf{q}^{j},\mathbf{q}^{k})}{\partial \mathbf{q}^{*}}\\
\frac{\partial \mathrm{f} (\mathbf{q},\mathbf{q}^{i},\mathbf{q}^{j},\mathbf{q}^{k})}{\partial \mathbf{q}^{i*}}\\
\frac{\partial \mathrm{f} (\mathbf{q},\mathbf{q}^{i},\mathbf{q}^{j},\mathbf{q}^{k})}{\partial \mathbf{q}^{j*}}\\
\frac{\partial \mathrm{f} (\mathbf{q},\mathbf{q}^{i},\mathbf{q}^{j},\mathbf{q}^{k})}{\partial \mathbf{q}^{k*}}
\end{bmatrix}=\frac{1}{4}\mathbf{A}\begin{bmatrix}
\frac{\partial \mathrm{f} (\mathbf{q}_{r},\mathbf{q}_{i},\mathbf{q}_{j},\mathbf{q}_{k})}{\partial \mathbf{q}_{r}}\\
\frac{\partial \mathrm{f} (\mathbf{q}_{r},\mathbf{q}_{i},\mathbf{q}_{j},\mathbf{q}_{k})}{\partial \mathbf{q}_{i}}\\
\frac{\partial \mathrm{f} (\mathbf{q}_{r},\mathbf{q}_{i},\mathbf{q}_{j},\mathbf{q}_{k})}{\partial \mathbf{q}_{j}}\\
\frac{\partial \mathrm{f} (\mathbf{q}_{r},\mathbf{q}_{i},\mathbf{q}_{j},\mathbf{q}_{k})}{\partial \mathbf{q}_{k}}
\end{bmatrix}\hspace{0.06cm}\text{that is}\hspace{0.06cm}\nabla_{\mathbf{q}^{a*}}\mathrm{f}=\frac{1}{4}\mathbf{A}\nabla\mathrm{f}
\label{eq:GradientMapping}
\end{equation}
where for brevity we use $\nabla\mathrm{f}$ to denote the gradient of $\mathrm{f}(\cdot)$ with respect to $\left[\mathbf{q}_{r},\mathbf{q}_{\imath},\mathbf{q}_{\jmath},\mathbf{q}_{\kappa}\right]$.

\subsection{Multiplication, Rotation, and Conjugation Rules in the $\mathbb{HR}$-Calculus}

\label{Sec:MRule}

Consider $\nu\in\mathbb{H}$ and $\xi\in\{1,\imath,\jmath,\kappa\}$. On one hand, multiplying both sides of \eqref{eq:HR-derivatives} by $\nu$ gives $\frac{\partial\mathrm{f}\left(\mathbf{q}^{a}\right)\nu}{\partial q^{\xi}}=\frac{\partial\mathrm{f}\left(\mathbf{q}^{a}\right)}{\partial q^{\xi}}\nu
$. On the  other hand, we have~\cite{GenHR}
\begin{equation}
\forall\nu\in\mathbb{H}\hspace{0.12cm}\&\hspace{0.12cm}\xi\in\{\imath,\jmath,\kappa\}:\hspace{0.12cm}\frac{\partial\nu\mathrm{f}\left(\mathbf{q}^{a}\right)}{\partial q^{\xi}}=\nu\frac{\partial\mathrm{f}\left(\mathbf{q}^{a}\right)}{\partial q^{\nu^{-1}\xi}}\cdot
\label{eq:MultiplicaationRule}
\end{equation}
Detailed proofs are given in~\cite{GenHR}; however, as a demonstrative example, from \eqref{eq:DiffQuaternion}, we have  
\begin{equation}
\begin{aligned}
\frac{\partial\nu\mathrm{f}}{\partial \mathbf{q}^{*}}=&\frac{1}{4}\left(\frac{\partial\nu\mathrm{f}}{\partial\mathbf{q}_{r}}+\imath\frac{\partial\nu\mathrm{f}}{\partial\nu\mathbf{q}_{\imath}}+\jmath\frac{\partial\nu\mathrm{f}}{\partial\mathbf{q}_{\jmath}}+\kappa\frac{\partial\nu \mathrm{f}}{\partial\mathbf{q}_{\kappa}}\right)
\\
=&\frac{1}{4}\left(\nu\frac{\partial\mathrm{f}}{\partial\mathbf{q}_{r}}+\nu\imath^{\nu^{-1}}\frac{\partial\mathrm{f}}{\partial\mathbf{q}_{\imath}}+\nu\jmath^{\nu^{-1}}\frac{\partial\mathrm{f}}{\partial\mathbf{q}_{\jmath}}+\nu\kappa^{\nu^{-1}}\frac{\partial \mathrm{f}}{\partial\mathbf{q}_{\kappa}}\right)=\nu\frac{\partial\mathrm{f}}{\partial \mathbf{q}^{\nu^{-1}*}}
\end{aligned}
\label{eq:Important}
\end{equation}
where we have used $\nu\xi^{\nu^{-1}}=\nu\nu^{-1}\xi\nu=\xi\nu$. In addition, note that $q^{\nu^{-1}}=q_{r}+\imath^{\nu^{-1}}q_{\imath}+\jmath^{\nu^{-1}}q_{\jmath}+\kappa^{\nu^{-1}}q_{\kappa}$ (see \eqref{eq:Involution}); hence, the final expression in \eqref{eq:Important} is the derivate given in \eqref{eq:DiffQuaternion} with its basis  transformed from $\{1,\imath,\jmath,\kappa\}$ to $\{1,\imath^{\nu^{-1}},\jmath^{\nu^{-1}},\kappa^{\nu^{-1}}\}$. 

Now, the effect of a rotation can be formulated via multiplication from both sides so that 
\begin{equation}
\begin{aligned}
\left(\frac{\partial\mathrm{f}\left(\mathbf{q}^{a}\right)}{\partial q^{\xi}}\right)^{\nu}=&\nu\frac{\partial\mathrm{f}\left(\mathbf{q}^{a}\right)}{\partial q^{\xi}}\nu^{-1}=\frac{\partial\nu\mathrm{f}\left(\mathbf{q}^{a}\right)}{\partial q^{\nu\xi}}\nu^{-1}\hspace{0.32cm}\text{(performed using \eqref{eq:MultiplicaationRule})}
\\
=&\frac{\partial\nu\mathrm{f}\left(\mathbf{q}^{a}\right)\nu^{-1}}{\partial q^{\nu\xi}}=\frac{\partial\mathrm{f}^{\nu}\left(\mathbf{q}^{a}\right)}{\partial q^{\nu\xi}}\cdot
\hspace{0.52cm}(\text{performed via the definition in \eqref{eq:Involution}})
\end{aligned}
\label{eq:RotationRule}
\end{equation}
In case of conjugation, we have
\begin{align}
\left(\frac{\partial\mathrm{f}\left(\mathbf{q}^{a}\right)}{\partial \mathbf{q}^{*}}\right)^{*}=&\frac{1}{4}\left(\frac{\partial\mathrm{f}^{*}\left(\mathbf{q}^{a}\right)}{\partial\mathbf{q}_{r}}-\frac{\partial\mathrm{f}^{*}\left(\mathbf{q}^{a}\right)}{\partial\mathbf{q}_{\imath}}\imath-\frac{\partial\mathrm{f}^{*}\left(\mathbf{q}^{a}\right)}{\partial\mathbf{q}_{\jmath}}\jmath-\frac{\partial \mathrm{f}^{*}\left(\mathbf{q}^{a}\right)}{\partial\mathbf{q}_{\kappa}}\kappa\right)\hspace{0.32cm}(\text{performed using \eqref{eq:RevMulConj}})\nonumber
\\
=&\frac{\partial_{\text{left}}\mathrm{f}^{*}\left(\mathbf{q}^{a}\right)}{\partial\mathbf{q}}\label{eq:ConjugagteRule}
\end{align}
where $ \partial_{\text{left}}$ denotes the placement of partial differential on the left side of the imaginary units $\{\imath,\jmath,\kappa\}$. This also follows from \eqref{eq:GradientMapping} which yields $
\left(\nabla_{\mathbf{q}^{a*}}\mathrm{f}\right)^{\H}=\frac{1}{4}\left(\nabla\mathrm{f}\right)^{\T}\mathbf{A}^{\H}$. These results can now be used to establish a chain derivative rule in the $\mathbb{HR}$-calculus. 

\begin{rem}
From \eqref{eq:ConjugagteRule}, an equivalent calculus can be derived via replacing $\partial$ with $\partial_{\text{left}}$, see~\cite{GenHR}. Although both approaches yield equivalent results, a skilful use of both approaches can simplify derived algorithms, see Section~\ref{Sec:QLMS}.
\end{rem}

Finally, consider $\{\mathrm{f}\left(\cdot\right),\mathrm{g}\left(\cdot\right)\}:\mathbb{H}^{M}\rightarrow\mathbb{H}$; then, $\mathrm{h}\left(\mathbf{q}\right)=\mathrm{f}\left(\mathbf{q}\right)\mathrm{g}\left(\mathbf{q}\right)$ can be expressed in terms of the real-valued components of $\mathrm{f}\left(\cdot\right)$ and $\mathrm{g}\left(\cdot\right)$ as
\begin{equation}
\mathrm{h}\left(\mathbf{q}\right)=\left(\Re\left\{\mathrm{f}\left(\mathbf{q}\right)\right\}+\dots+\Im_{\kappa}\left\{\mathrm{f}\left(\mathbf{q}\right)\right\}\right)\left(\Re\left\{\mathrm{g}\left(\mathbf{q}\right)\right\}+\dots+\Im_{\kappa}\left\{\mathrm{g}\left(\mathbf{q}\right)\right\}\right)
\label{eq:TheThing}
\end{equation}
with the exact expression given in SM-3. Now, $\forall\xi\in\{1,\imath,\jmath,\kappa\}$, $\frac{\partial\mathrm{h}\left(\mathbf{q}\right)}{\partial\mathbf{q}^{\xi}}$ can be related to the derivatives of $\mathrm{h}\left(\mathbf{q}\right)$ with respect to $\{\mathbf{q}_{r},\mathbf{q}_{\imath},\mathbf{q}_{\jmath},\mathbf{q}_{\kappa}\}$ in manner similar to the framework in \eqref{eq:AugmentedFunction}-\eqref{eq:GradientMapping}. In this setting, after long mathematical manipulations that are omitted (but can be found in~\cite{GenHR,Cyrus}), we arrive at\footnote{In a similar fashion it can be shown that $\frac{\partial\mathrm{h}\left(\mathbf{q}\right)}{\partial\mathbf{q}^{\xi}}=\mathrm{f}\left(\mathbf{q}\right)\frac{\partial\mathrm{g}\left(\mathbf{q}\right)}{\partial\mathbf{q}^{\mathrm{f}^{*}\left(\mathbf{q}\right)\xi}}+\frac{\partial\mathrm{f}\left(\mathbf{q}\right)}{\partial\mathbf{q}^{\xi}}\mathrm{g}\left(\mathbf{q}\right)$~\cite{GenHR}.}

\begin{equation}
\forall\xi\in\{1,\imath,\jmath,\kappa\}:
\frac{\partial\mathrm{h}\left(\mathbf{q}\right)}{\partial\mathbf{q}^{\xi*}}=\mathrm{f}\left(\mathbf{q}\right)\frac{\partial\mathrm{g}\left(\mathbf{q}\right)}{\partial\left(\mathbf{q}^{\mathrm{f}^{*}\left(\mathbf{q}\right)\xi}\right)^{*}}+\frac{\partial\mathrm{f}\left(\mathbf{q}\right)}{\partial\mathbf{q}^{\xi*}}\mathrm{g}\left(\mathbf{q}\right).
\label{eq:MultiplicaationRuleFunction}
\end{equation}

\begin{rem}
This notion of gradient accommodates the derivation of Taylor series expansions for quaternion-valued function. This is presented in SM-2.
\end{rem}

\subsection{Chain Derivative Rule in the $\mathbb{HR}$-Calculus}

\label{Sec:ChainRule}

Consider the quaternion-valued compound function $\mathrm{f}\left(\mathrm{g}\left(\cdot\right)\right):\mathbb{H}^{M}\rightarrow\mathbb{H}$, where $\mathrm{f}\left(\cdot\right):\mathbb{H}^{M}\rightarrow\mathbb{H}$ and $\mathrm{g}\left(\cdot\right):\mathbb{H}^{M}\rightarrow\mathbb{H}^{M}$. One can consider the relation between $\mathrm{g}^{a}\left(\mathbf{q}\right)$ and the real-valued components of $\mathrm{g}\left(\mathbf{q}\right)$, akin to that given in \eqref{eq:MappingHR} and \eqref{eq:HR-derivatives}, yielding
\begin{equation}
\begin{bmatrix}
\frac{\partial\mathrm{f}\left(\mathrm{g}^{a}\left(\mathbf{q}\right)\right)}{\partial\mathrm{g}\left(\mathbf{q}\right)}\\
\frac{\partial\mathrm{f}\left(\mathrm{g}^{a}\left(\mathbf{q}\right)\right)}{\partial\mathrm{g}^{\imath}\left(\mathbf{q}\right)}\\
\frac{\partial\mathrm{f}\left(\mathrm{g}^{a}\left(\mathbf{q}\right)\right)}{\partial \mathrm{g}^{\jmath}\left(\mathbf{q}\right)}\\
\frac{\partial\mathrm{f}\left(\mathrm{g}^{a}\left(\mathbf{q}\right)\right)}{\partial \mathrm{g}^{\kappa}\left(\mathbf{q}\right)}
\end{bmatrix}
=\frac{1}{4}\mathbf{A}^{*}
\begin{bmatrix}
\frac{\partial\mathrm{f}\left(\mathrm{g}^{a}\left(\mathbf{q}\right)\right)}{\partial\mathrm{g}_{r}\left(\mathbf{q}\right)}\\
\frac{\partial\mathrm{f}\left(\mathrm{g}^{a}\left(\mathbf{q}\right)\right)}{\partial\mathrm{g}_{\imath}\left(\mathbf{q}\right)}
\\
\frac{\partial\mathrm{f}\left(\mathrm{g}^{a}\left(\mathbf{q}\right)\right)}{\partial\mathrm{g}_{\jmath}\left(\mathbf{q}\right)}
\\
\frac{\partial\mathrm{f}\left(\mathrm{g}^{a}\left(\mathbf{q}\right)\right)}{\partial\mathrm{g}_{\kappa}\left(\mathbf{q}\right)}
\end{bmatrix}\hspace{0.26cm}\text{with}\hspace{0.26cm}\begin{bmatrix}
\frac{\partial\mathrm{g}_{r}\left(\mathbf{q}\right)}{\partial\mathbf{q}}
\\
\frac{\partial\mathrm{g}_{\imath}\left(\mathbf{q}\right)}{\partial\mathbf{q}}
\\
\frac{\partial\mathrm{g}_{\jmath}\left(\mathbf{q}\right)}{\partial\mathbf{q}}
\\
\frac{\partial\mathrm{g}_{\kappa}\left(\mathbf{q}\right)}{\partial\mathbf{q}}
\end{bmatrix}=\frac{1}{4}\mathbf{A}^{\H}\begin{bmatrix}
\frac{\partial\mathrm{g}\left(\mathbf{q}\right)}{\partial\mathbf{q}}
\\
\frac{\partial\mathrm{g}^{\imath}\left(\mathbf{q}\right)}{\partial\mathbf{q}}
\\
\frac{\partial\mathrm{g}^{\jmath}\left(\mathbf{q}\right)}{\partial\mathbf{q}}\\\frac{\partial\mathrm{g}^{\kappa}\left(\mathbf{q}\right)}{\partial\mathbf{q}}
\end{bmatrix}\cdot
\label{eq:ChainSet}
\end{equation}
Following the framework set in \eqref{eq:HR-derivatives}-\eqref{eq:GradientMapping} gives
\begin{equation}
\begin{aligned}
\forall\xi\in\{\imath,\jmath,\kappa\}:\hspace{0.12cm}\frac{\partial\mathrm{f}\left(\mathrm{g}\left(\mathbf{q}\right)\right)}{\partial\mathbf{q}^{\xi}}=&\frac{\partial\mathrm{g}_{r}\left(\mathbf{q}\right)}{\partial\mathbf{q}^{\xi}}\frac{\partial\mathrm{f}\left(\mathrm{g}\left(\mathbf{q}\right)\right)}{\partial\mathrm{g}_{r}\left(\mathbf{q}\right)}+\frac{\partial\mathrm{g}_{\imath}\left(\mathbf{q}\right)}{\partial\mathbf{q}^{\xi}}\frac{\partial\mathrm{f}\left(\mathrm{g}\left(\mathbf{q}\right)\right)}{\partial\mathrm{g}_{\imath}\left(\mathbf{q}\right)}
\\
&+\frac{\partial\mathrm{g}_{\jmath}\left(\mathbf{q}\right)}{\partial\mathbf{q}^{\xi}}\frac{\partial\mathrm{f}\left(\mathrm{g}\left(\mathbf{q}\right)\right)}{\partial\mathrm{g}_{\jmath}\left(\mathbf{q}\right)}+\frac{\partial\mathrm{g}_{\kappa}\left(\mathbf{q}\right)}{\partial\mathbf{q}^{\xi}}\frac{\partial\mathrm{f}\left(\mathrm{g}\left(\mathbf{q}\right)\right)}{\partial\mathrm{g}_{\kappa}\left(\mathbf{q}\right)}\cdot
\end{aligned}
\label{eq:ChainMid}
\end{equation}
Upon substituting \eqref{eq:ChainSet} into \eqref{eq:ChainMid} we arrive at 
\begin{equation}
\begin{aligned}
\forall\xi\in\{\imath,\jmath,\kappa\}:\hspace{0.12cm}\frac{\partial\mathrm{f}\left(\mathrm{g}\left(\mathbf{q}\right)\right)}{\partial\mathbf{q}^{\xi}}=&\frac{\partial\mathrm{g}\left(\mathbf{q}\right)}{\partial\mathbf{q}^{\xi}}\frac{\partial\mathrm{f}\left(\mathrm{g}\left(\mathbf{q}\right)\right)}{\partial\mathrm{g}\left(\mathbf{q}\right)}+\frac{\partial\mathrm{g}^{\imath}\left(\mathbf{q}\right)}{\partial\mathbf{q}^{\xi}}\frac{\partial\mathrm{f}\left(\mathrm{g}\left(\mathbf{q}\right)\right)}{\partial\mathrm{g}^{\imath}\left(\mathbf{q}\right)}
\\
&+\frac{\partial\mathrm{g}^{\jmath}\left(\mathbf{q}\right)}{\partial\mathbf{q}^{\xi}}\frac{\partial\mathrm{f}\left(\mathrm{g}\left(\mathbf{q}\right)\right)}{\partial\mathrm{g}^{\jmath}\left(\mathbf{q}^{\xi}\right)}+\frac{\partial\mathrm{g}^{\kappa}\left(\mathbf{q}\right)}{\partial\mathbf{q}^{\xi}}\frac{\partial\mathrm{f}\left(\mathrm{g}\left(\mathbf{q}\right)\right)}{\partial\mathrm{g}^{\kappa}\left(\mathbf{q}\right)}
\end{aligned}
\label{eq:ChainMidMid}
\end{equation}
with the direction of steepest change given by
\begin{equation}
\begin{aligned}
\frac{\partial\mathrm{f}\left(\mathrm{g}\left(\mathbf{q}\right)\right)}{\partial\mathbf{q}^{*}}=&\frac{\partial\mathrm{g}^{*}\left(\mathbf{q}\right)}{\partial\mathbf{q}^{*}}\frac{\partial\mathrm{f}\left(\mathrm{g}\left(\mathbf{q}\right)\right)}{\partial\mathrm{g}^{*}\left(\mathbf{q}\right)}+\frac{\partial\mathrm{g}^{\imath*}\left(\mathbf{q}\right)}{\partial\mathbf{q}^{*}}\frac{\partial\mathrm{f}\left(\mathrm{g}^{\imath}\left(\mathbf{q}\right)\right)}{\partial\mathrm{g}^{\imath*}\left(\mathbf{q}\right)}
\\
&+\frac{\partial\mathrm{g}^{\jmath*}\left(\mathbf{q}\right)}{\partial\mathbf{q}^{*}}\frac{\partial\mathrm{f}\left(\mathrm{g}^{\jmath}\left(\mathbf{q}\right)\right)}{\partial\mathrm{g}^{\jmath*}\left(\mathbf{q}\right)}+\frac{\partial\mathrm{g}^{\kappa*}\left(\mathbf{q}\right)}{\partial\mathbf{q}^{*}}\frac{\partial\mathrm{f}\left(\mathrm{g}^{\kappa}\left(\mathbf{q}\right)\right)}{\partial\mathrm{g}^{\kappa*}\left(\mathbf{q}\right)}\cdot
\end{aligned}
\label{eq:ChainMidMidMid}
\end{equation}
Most useful in adaptive information processing applications is the case where $\mathrm{g}\left(\cdot\right):\mathbb{H}^{M}\rightarrow\mathbb{R}$ and $\mathrm{f}\left(\cdot\right):\mathbb{R}\rightarrow\mathbb{R}$. In this case, from  \eqref{eq:ChainMid}-\eqref{eq:ChainMidMidMid}, we have
\begin{equation}
\forall\xi\in\{\imath,\jmath,\kappa\}:\hspace{0.12cm}\frac{\partial\mathrm{f}\left(\mathrm{g}\left(\mathbf{q}\right)\right)}{\partial\mathbf{q}^{\xi}}=\frac{\partial\mathrm{g}\left(\mathbf{q}\right)}{\partial\mathbf{q}^{\xi}}\frac{\partial\mathrm{f}\left(\mathrm{g}\left(\mathbf{q}\right)\right)}{\partial\mathrm{g}\left(\mathbf{q}\right)}\hspace{0.12cm}\text{and}\hspace{0.12cm}\frac{\partial\mathrm{f}\left(\mathrm{g}\left(\mathbf{q}\right)\right)}{\partial\mathbf{q}^{\xi*}}=\frac{\partial\mathrm{g}\left(\mathbf{q}\right)}{\partial\mathbf{q}^{\xi*}}\frac{\partial\mathrm{f}\left(\mathrm{g}\left(\mathbf{q}\right)\right)}{\partial\mathrm{g}\left(\mathbf{q}\right)}\cdot
\label{eq:RealChain}
\end{equation}
The chain derivative rule allows for the derivation of nonlinear quaternion-valued adaptive learning techniques. The case of linear (\textit{cf}. nonlinear) adaptive filters is presented in Section~\ref{Sec:QLMS} (\textit{cf}. SM-4), while its application in quaternion-valued backpropagation is demonstrated in SM-5.

\section{Information Processing and Decision Making}

Thus far, the aim has been to present the reader with formulation of useful $\mathbb{HR}$-calculus tools, while outlining their derivation philosophy. In this section, the introduced $\mathbb{HR}$-calculus tools are used to present a number of widely used adaptive information processing and decision making techniques, that are formulated directly in the quaternion formulation.

\subsection{Quaternion-Valued Least Mean Square (QLMS)}

\label{Sec:QLMS}
	
Given the most elementary learning task, the aim is to learn weight vector $\mathbf{w}_{\left[n\right]}$ so that for input sequence $\{\mathbf{z}_{[n]}:n\in\mathbb{N}\}$, $\hat{y}_{n}=\mathbf{w}^{\T}_{\left[n\right]}\mathbf{z}^{a}_{\left[n\right]}$ tracks observation sequence $\{y_{\left[n\right]}:n\in\mathbb{N}\}$ in a manner that minimises the cost function
\begin{equation}
J_{\left[n\right]}=\|\epsilon_{\left[n\right]}\|^2\hspace{0.26cm}\text{with}\hspace{0.26cm}\epsilon_{\left[n\right]}=y_{\left[n\right]}-\hat{y}_{\left[n\right]}.
\end{equation}
This can be achieved via iterative updates of the weight vector as
\begin{equation}
\mathbf{w}_{\left[n+1\right]}=\mathbf{w}_{\left[n\right]}-\gamma\nabla_{\mathbf{w}^{*}}J_{\left[n\right]}
\end{equation}
where $\gamma\in\mathbb{R}^{+}$ is an adaptation gain. Now, considering that $\|\epsilon_{\left[n\right]}\|\in\mathbb{R}$ and using the chain derivative rule of the $\mathbb{HR}$-calculus given in \eqref{eq:RealChain}, we have
\begin{equation}
\nabla_{\mathbf{w}^{*}}J_{\left[n\right]}=\frac{\partial\|\epsilon_{\left[n\right]}\|^{2}}{\partial\|\epsilon_{\left[n\right]}\|}\nabla_{\mathbf{w}^{*}}\|\epsilon_{\left[n\right]}\|=-\frac{1}{2}\epsilon_{\left[n\right]}\mathbf{z}^{a*}_{\left[n\right]}.
\end{equation}
Thus, the weight update term becomes
\begin{equation}
\mathbf{w}_{\left[n+1\right]}=\mathbf{w}_{\left[n\right]}+\gamma\epsilon_{\left[n\right]}\mathbf{z}^{a*}_{\left[n\right]}
\label{eq:GradLeft}
\end{equation}
with the $\frac{1}{2}$ factor absorbed into the adaptation gain.  Alternatively, using the multiplication rule in \eqref{eq:MultiplicaationRuleFunction}, we can also have
\begin{equation}
\begin{aligned}
\nabla_{\mathbf{w}^{*}}J_{\left[n\right]}=&\nabla_{\mathbf{w}^{*}}\|\epsilon_{\left[n\right]}\|^{2}=\nabla_{\mathbf{w}^{*}}\left(\epsilon^{*}_{\left[n\right]}\epsilon_{\left[n\right]}\right)=\epsilon^{*}_{\left[n\right]}\left(\nabla_{\mathbf{w}^{*}}\epsilon_{\left[n\right]}\right)+\left(\nabla_{\mathbf{w}^{\epsilon_{\left[n\right]*}}}\epsilon^{*}_{\left[n\right]}\right)\epsilon_{\left[n\right]}
\\
=&\frac{1}{2}\epsilon^{*}_{\left[n\right]}\mathbf{z}^{a*}_{\left[n\right]}-\mathbf{z}^{a*}_{\left[n\right]}\Re\left\{\epsilon_{\left[n\right]}\right\}=-\frac{1}{2}\epsilon_{\left[n\right]}\mathbf{z}^{a*}_{\left[n\right]}.
\end{aligned}
\label{eq:GradRight}
\end{equation}
There are different derivations available, e.g. see~\cite{GenHR,Cyrus,HR-Gradient}. Although these derivations might seem different from those in \eqref{eq:GradLeft} and \eqref{eq:GradRight}, they operate in the same manner.  

Finally, if we assume that $\mathbf{w}^{\text{opt}}$ exists so that $y_{\left[n\right]}=\mathbf{w}^{\text{opt}^{\T}}\mathbf{z}^{a}_{\left[n\right]}\hspace{0.26cm}\text{and}\hspace{0.26cm}\boldsymbol{\varepsilon}_{\left[n\right]}=\mathbf{w}^{\text{opt}}-\mathbf{w}_{\left[n\right]}$. Then, from \eqref{eq:GradLeft} it follows that $\boldsymbol{\varepsilon}^{\H}_{\left[n+1\right]}=\boldsymbol{\varepsilon}^{\H}_{\left[n\right]}-\gamma\mathbf{z}^{a}_{\left[n\right]}\mathbf{z}^{a\H}_{\left[n\right]}\boldsymbol{\varepsilon}^{\H}_{\left[n\right]}$ which in turn yields the quaternion-valued Lyapunov function
\begin{equation}
\boldsymbol{\Sigma}_{\boldsymbol{\varepsilon}^{a}_{\left[n+1\right]}}=\left(\mathbf{I}-\gamma\mathbf{G}_{\left[n\right]}\right)\boldsymbol{\Sigma}_{\boldsymbol{\varepsilon}_{\left[n\right]}}\left(\mathbf{I}-\gamma\mathbf{G}_{\left[n\right]}\right)^{\H}
\end{equation}
where $\boldsymbol{\Sigma}_{\boldsymbol{\varepsilon}^{a}_{\left[n\right]}}=\E{\boldsymbol{\varepsilon}^{a}_{\left[n\right]}\boldsymbol{\varepsilon}^{a\H}_{\left[n\right]}}$, $\boldsymbol{\Sigma}_{\mathbf{z}^{a}_{\left[n\right]}}=\E{\mathbf{z}^{a}_{\left[n\right]}\mathbf{z}^{a\H}_{\left[n\right]}}$, and $\mathbf{G}_{\left[n\right]}=\text{diag}\left\{\boldsymbol{\Sigma}_{\mathbf{z}^{a}_{\left[n\right]}},\boldsymbol{\Sigma}^{\imath}_{\mathbf{z}^{a}_{\left[n\right]}},\boldsymbol{\Sigma}^{\jmath}_{\mathbf{z}^{a}_{\left[n\right]}},\boldsymbol{\Sigma}^{\kappa}_{\mathbf{z}^{a}_{\left[n\right]}}\right\}$. Thus, if $\gamma<\p{\mathbf{G}_{\left[n\right]}}$ and $\boldsymbol{\Sigma}_{\mathbf{z}^{a}_{\left[n\right]}}$ is full rank; then, as $n\rightarrow\infty$ one has $\p{\boldsymbol{\Sigma}_{\mathbf{z}^{a}_{\left[n\right]}}}\rightarrow0$, resulting in a convergent filtering operation.

\begin{figure}[h!]
\centering
\includegraphics[width=0.65\linewidth,trim= 0cm 0cm 0cm 0.86cm]{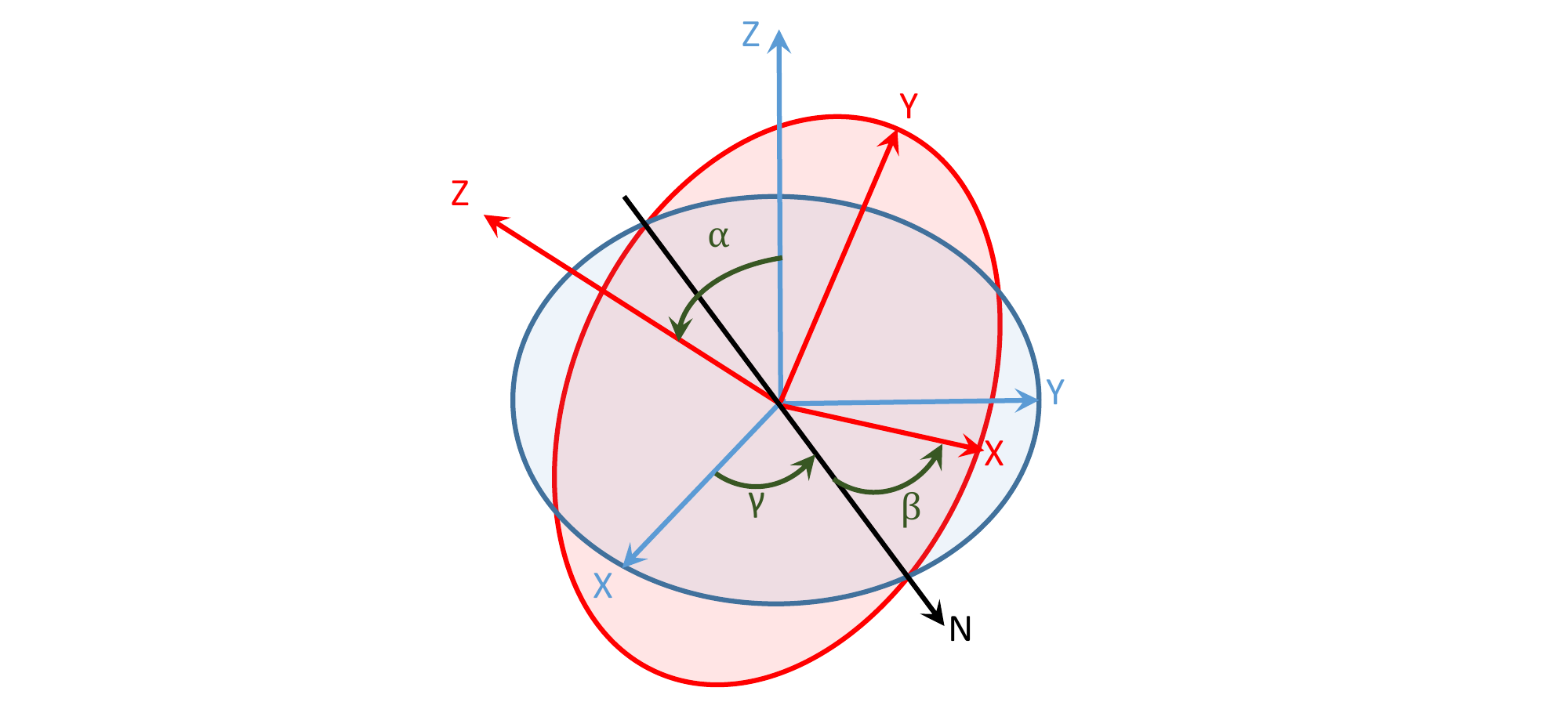}
\caption{Setting of the inertial body motion sensor with the fixed coordinate system (blue), sensor coordinate system (red), and Euler angles (green). The ``N'' axis is used as a visual guide to indicate the yaw angle. Figure taken from~\cite{PouriaPhD}.}
\label{Fig:EulerAngles}
\end{figure}

Now, let us present the application of QLMS in body motion tracking. Modern motion sensors record the Euler angles, as shown in Fig.~\ref{Fig:EulerAngles}.  In order to make this data suitable for processing each angle is converted into a quaternion using the following transformations~\cite{QRKHS}
\begin{equation}
\alpha\rightarrow e^{\imath\frac{\alpha}{2}},\hspace{0.12cm}\beta\rightarrow e^{\jmath\frac{\beta}{2}},\hspace{0.12cm}\gamma\rightarrow e^{\kappa\frac{\gamma}{2}}
\label{eq:first transform}
\end{equation}
with each term respectively representing the roll, pitch, and yaw rotations~\cite{Kuipers,PouriaPhD}. The three quaternion terms in \eqref{eq:first transform} are then combined into one quaternion-valued variable given by~\cite{PouriaPhD}
\begin{equation}
q=\big(e^{\kappa\frac{\gamma}{2}}\big)\big(e^{\jmath\frac{\beta}{2}}\big)\big(e^{\imath\frac{\alpha}{2}}\big)
\label{eq:final transform}
\end{equation}
representing the overall rotation.

\begin{figure}[!h]
\centering
\includegraphics[width=0.9\linewidth,trim = 0cm 0.36cm 0cm 0cm]{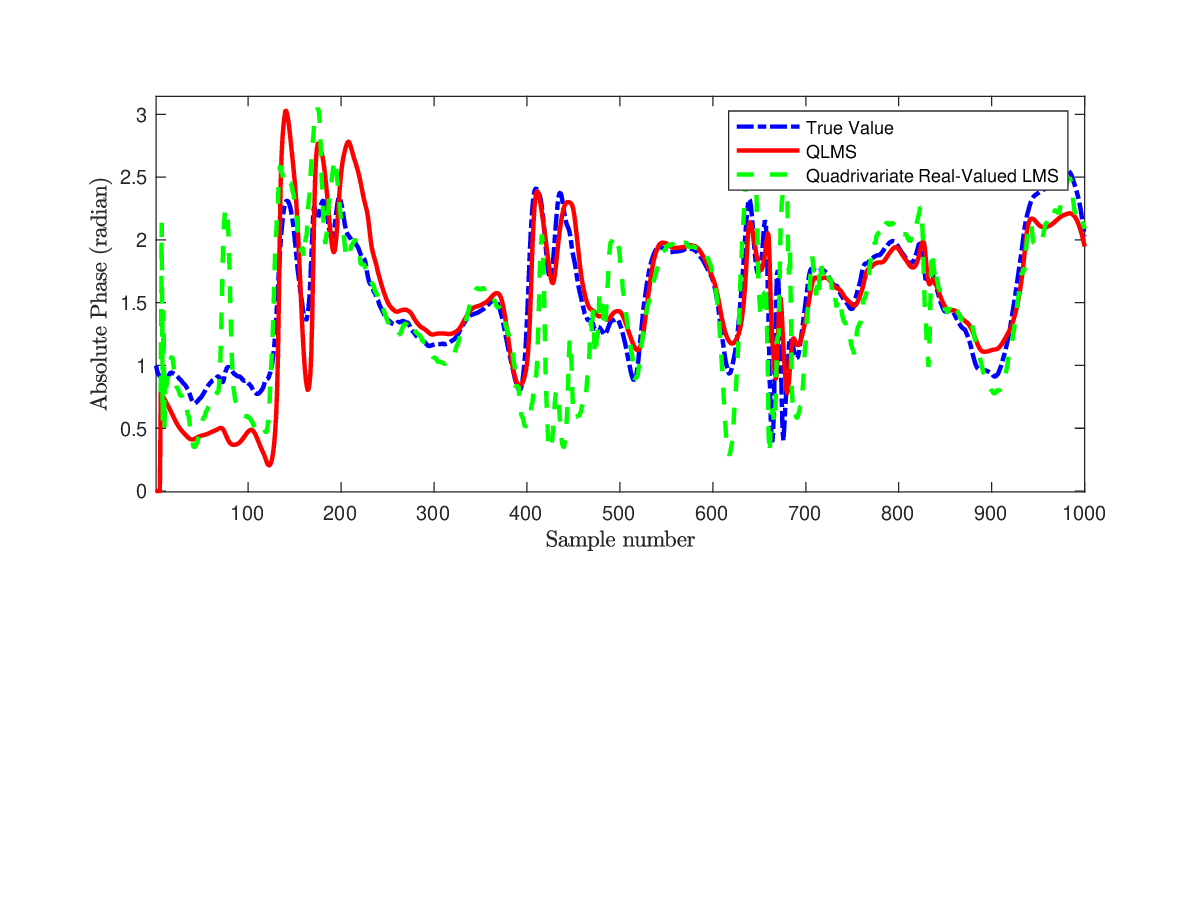}
\caption{Absolute value of the phase of quaternion-valued body motion signal employing the QLMS and a quadrivariate real-valued LMS approach.}
\label{fig:abs of first samples}
\end{figure}

As an illustrative example, we recover the experiment in~\cite{PouriaPhD} using the frameworks in this article.  The QLMS was considered in a one-step-ahead prediction setting to track body rotations. Fig.~\ref{fig:abs of first samples} shows the estimates of the absolute value of the phase for the body motion signal obtained by the QLMS filter and its real-valued quadrivariate counterpart, that is, the case where the quaternion-valued body motion signal was considered as a four-element vector with the real-valued least mean square (LMS) algorithm used to track the signal. Notice that the quaternion-valued  filter had a significantly better performance than the quadrivariate LMS algorithm. This is mainly attributable to the existence of a natural notion of phase angle in the quaternion domain, whereas using real-valued algebras requires extra operation to formulate phase angles, which are overly sensitive to finite estimation accuracy of their inputs. 

\subsection{The Quaternion Kalman Filter}

\label{Sec:Kalman}

The aim is to track the state vector sequence $\{\mathbf{x}_{\left[n\right]}: n\in\mathbb{N}\}$ representing the sate of a physical system and formulated as
\begin{equation}
\mathbf{x}^{a}_{\left[n+1\right]}=\mathbf{A}\mathbf{x}^{a}_{\left[n\right]}+\mathbf{v}^{a}_{\left[n\right]}
\label{eq:State-Model}
\end{equation}
where $\mathbf{A}$ and $\mathbf{v}_{\left[n\right]}$ are the evolution function and a noise vector at time instant $n$. Informative observations regarding the state vector are assumed available through observations
\begin{equation}
\mathbf{y}^{a}_{\left[n\right]}=\mathbf{H}\mathbf{x}^{a}_{\left[n\right]}+\mathbf{w}^{a}_{\left[n\right]}
\label{eq:Observation-Model}
\end{equation}
where $\mathbf{H}$ denotes the observation function, while  $\mathbf{y}_{\left[n\right]}$ and $\mathbf{w}_{\left[n\right]}$ are the observation and observation noise at time instant $n$. The aim is to formulate adaptation gain $\mathbf{G}_{[n]}$ so that the estimate 
\begin{equation}
\hat{\mathbf{x}}^{a}_{\left[n+1\right]}=\mathbf{A}\hat{\mathbf{x}}^{a}_{\left[n\right]}+\mathbf{G}_{\left[n\right]}\left(\mathbf{y}_{\left[n\right]}-\mathbf{H}\mathbf{A}\hat{\mathbf{x}}^{a}_{\left[n\right]}\right)
\end{equation}
minimises the cost function $\left\|\boldsymbol{\epsilon}_{\left[n+1\right]}\right\|^{2}=\frac{1}{4}\tr{\boldsymbol{\epsilon}^{a}_{\left[n+1\right]}\boldsymbol{\epsilon}^{a\H}_{\left[n+1\right]}}$ where
\begin{equation}
\boldsymbol{\epsilon}^{a}_{\left[n+1\right]}=\mathbf{x}^{a}_{[n]}-\hat{\mathbf{x}}^{a}_{[n]}=\left(\mathbf{I}-\mathbf{G}_{\left[n\right]}\mathbf{H}\right)\mathbf{A}\boldsymbol{\epsilon}^{a}_{\left[n\right]}+\left(\mathbf{I}-\mathbf{G}_{\left[n\right]}\mathbf{H}\right)\mathbf{v}^{a}_{\left[n\right]}+\mathbf{G}_{n}\mathbf{w}^{a}_{\left[n\right]}.
\label{eq:Error}
\end{equation}
From \eqref{eq:Error}, we have 
\begin{equation}
\begin{aligned}
\boldsymbol{\epsilon}^{a}_{\left[n+1\right]}\boldsymbol{\epsilon}^{a\H}_{\left[n+1\right]}=&\left(\mathbf{I}-\mathbf{G}_{\left[n\right]}\mathbf{H}\right)\mathbf{A}\boldsymbol{\epsilon}^{a}_{\left[n\right]}\boldsymbol{\epsilon}^{a\H}_{\left[n\right]}\mathbf{A}^{\H}\left(\mathbf{I}-\mathbf{G}_{\left[n\right]}\mathbf{H}\right)^{\H}
\\
&+\left(\mathbf{I}-\mathbf{G}_{\left[n\right]}\mathbf{H}\right)\mathbf{v}^{a}_{\left[n\right]}\mathbf{v}^{a\H}_{\left[n\right]}\left(\mathbf{I}-\mathbf{G}_{\left[n\right]}\mathbf{H}\right)^{\H}+\mathbf{G}_{n}\mathbf{w}^{a}_{\left[n\right]}\mathbf{w}^{a\H}_{\left[n\right]}\mathbf{G}^{\H}_{\left[n\right]}.
\end{aligned}
\label{eq:ErrorEvo1}
\end{equation}
Taking the statistical expectation of \eqref{eq:ErrorEvo1} and solving for $\partial\E{\left\|\boldsymbol{\epsilon}_{\left[n\right]}\right\|^{2}}/\partial\mathbf{G}_{\left[n\right]}=0$ yields
\begin{equation}
\mathbf{G}_{\left[n\right]}=\mathbf{M}_{\left[n\right]}\mathbf{H}^{\H}\boldsymbol{\Sigma}_{\mathbf{w}^{a}_{\left[n\right]}}^{-1}
\label{eq:Gain}
\end{equation}
where $\boldsymbol{\Sigma}_{\mathbf{w}^{a}_{\left[n\right]}}=\E{\mathbf{w}^{a}_{\left[n\right]}\mathbf{w}^{a\H}_{\left[n\right]}}$ and $\mathbf{M}_{[n]}=\E{\boldsymbol{\epsilon}^{a}_{\left[n\right]}\boldsymbol{\epsilon}^{a\H}_{\left[n\right]}}$ are positive semi-definite matrices. Furthermore, replacing \eqref{eq:Gain} into \eqref{eq:ErrorEvo1} gives the Riccati recursions
\begin{equation}
\mathbf{M}_{\left[n+1\right]}=\left(\left(\mathbf{A}\mathbf{M}_{\left[n\right]}\mathbf{A}^{\H}+\boldsymbol{\Sigma}_{\mathbf{v}^{a}_{\left[n\right]}}\right)^{-1}+\mathbf{H}^{\H}\boldsymbol{\Sigma}^{-1}_{\mathbf{w}^{a}_{\left[n\right]}}\mathbf{H}\right)^{-1}.
\label{eq:ErrorRiccati}
\end{equation}

\begin{figure}
\includegraphics[width=0.9\linewidth]{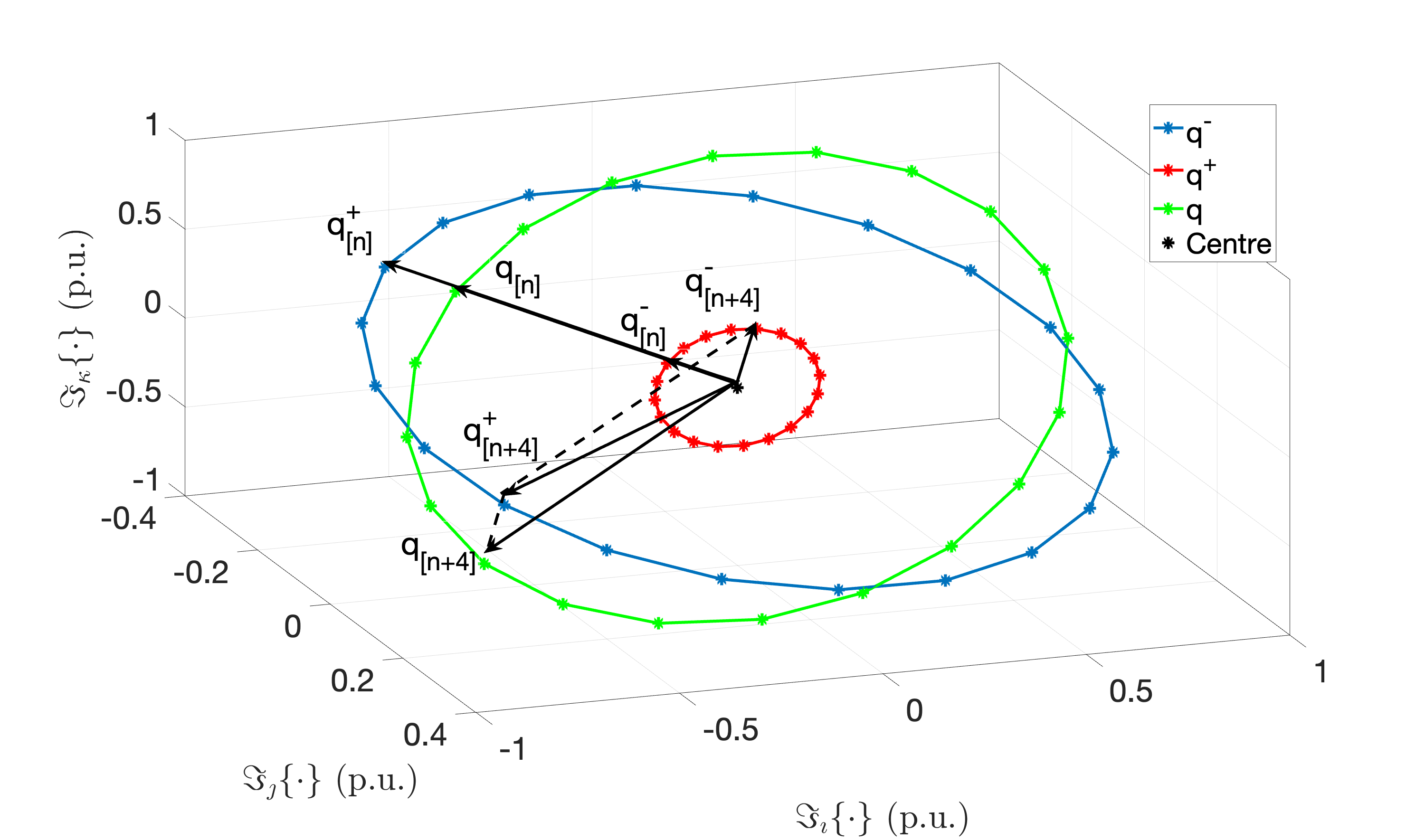}
\caption{System voltage, $q_{\left[n\right]}$, positive sequence element, $q^{+}_{\left[n\right]}$, and negative sequence element, $q^{-}_{\left[n\right]}$, of an unbalanced three-phase system suffering from an 80\% drop in the amplitude of one phase and $20$ degree shifts in the other phases. Note that p.u. stands for per unit, i.e. the variables are normalised to their nominal value.}
\label{Fig:Plot2}
\end{figure}

\begin{figure}
\centering
\includegraphics[width=0.9\linewidth, trim = 0cm 0.42cm 0cm 0cm]{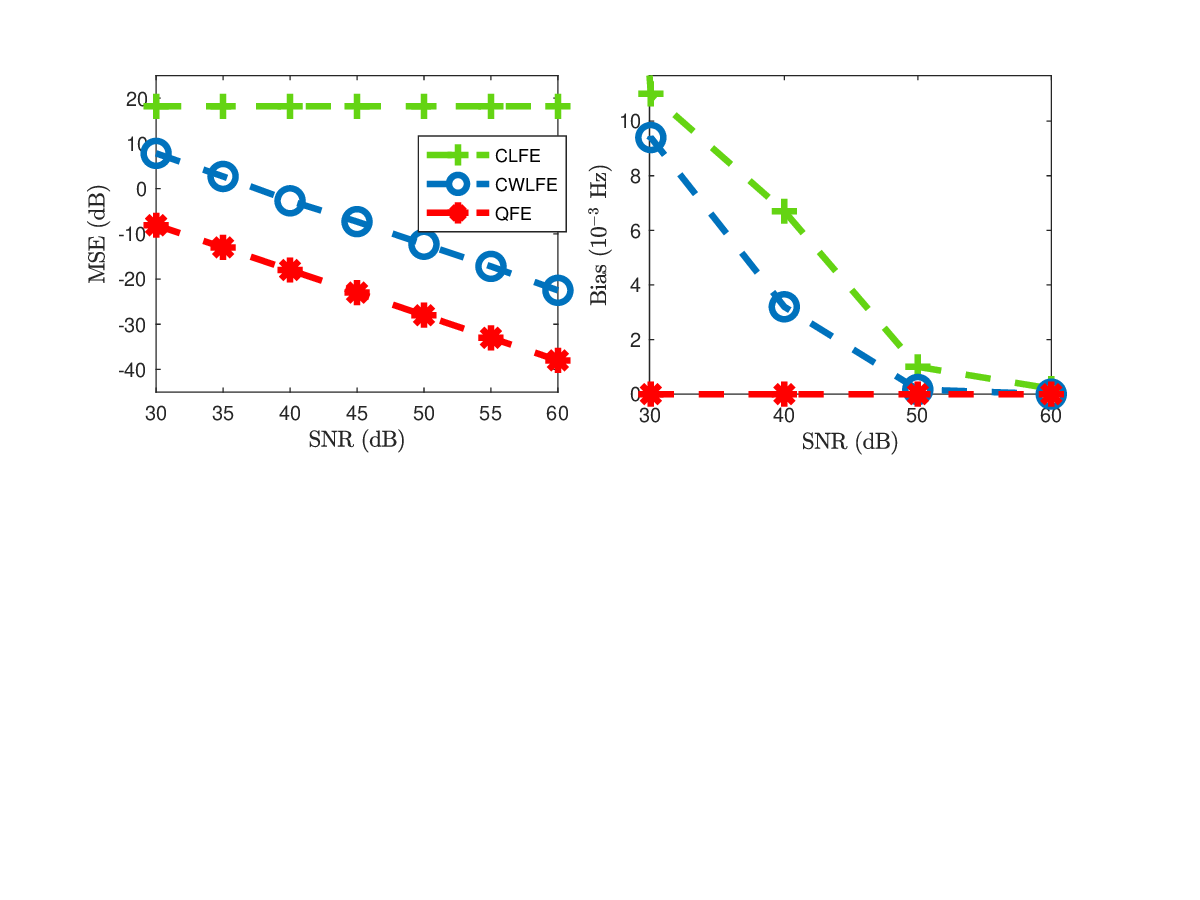}
\caption{Frequency estimation performance (bias and MSE) using the derived quaternion-valued model, quaternion frequency estimator (QFE), as compared to complex-valued linear (CLFE) and widely liner (CWLFE) techniques detailed in~\cite{PouriaPhD}. The results pertain to the unbalanced system in Figure~\ref{Fig:Plot2}.} 
\label{Fig:BiasMSE}
\end{figure}

One application example where quaternion Kalman filter has proven advantageous is frequency tracking in three-phase power systems. Instantaneous voltages of such systems can be formulate into the quaternion signal
\begin{equation}
\begin{aligned}
q_{\left[n\right]}=&\imath V_{a,\left[n\right]}\sin\left(2\pi f \Delta T n + \phi_{a,\left[n\right]}\right)+\jmath V_{b,\left[n\right]}\sin\left(2\pi f \Delta T n +  \phi_{b,\left[n\right]} +  \frac{2 \pi}{3}\right)
\\
&+\kappa V_{c,\left[n\right]}\sin\left(2\pi f \Delta T n +  \phi_{c,\left[n\right]} +  \frac{4 \pi}{3}\right)
\end{aligned}
\label{eq:PhaseVoltages}
\end{equation}
where $\{V_{a,\left[n\right]},V_{b,\left[n\right]},V_{c,\left[n\right]}\}$ (\textit{cf}. $\{\phi_{a,\left[n\right]},\phi_{b,\left[n\right]},\phi_{c,\left[n\right]}\}$) are the instantaneous amplitudes (\textit{cf}.  instantaneous phases) on phase $a$, $b$, and $c$, while $f$ is the system frequency\footnote{Frequency is common among all phases due to structure of power generation systems.} and $\Delta T$ is the sampling interval. Considering that all elements of \eqref{eq:PhaseVoltages} have the same frequency, it was shown in \cite{Phase3Freq} that $q_{\left[n\right]}$ traces an ellipse in a plane of the three-dimensional imaginary subspace of $\mathbb{H}$. Now, consider a new set of imaginary units, $\{\zeta, \zeta',\zeta''\}$ such that
\begin{equation}
\zeta \zeta' = \zeta'',\hspace{0.5em} \zeta' \zeta'' =\zeta,\hspace{0,5em} \zeta'' \zeta = \zeta'
\label{eq:new imaginary units}
\end{equation}
and $\zeta$ -$\zeta'$ reside in the same plane as $q_{\left[n\right]}$ (resulting in $\zeta''$ being normal to the plane). An arbitrary ellipse in the $\zeta-\zeta'$ plane can then be expressed as
\begin{equation}
q_{\left[n\right]}= \big(A_{\left[n\right]}\sin(2 \pi f \Delta T n + \phi_{\zeta,\left[n\right]})+ \zeta'' B_{\left[n\right]}\sin(2 \pi f \Delta T n + \phi_{\zeta',\left[n\right]})\big)\zeta.
\label{eq:SimpleForm}
\end{equation}  
where $A_{\left[n\right]},B_{\left[n\right]} \in \mathbb{R}$, are instantaneous amplitudes and $\phi_{\zeta,\left[n\right]},\phi_{\zeta',\left[n\right]}$ are instantaneous phases. Replacing $\text{sin}(\cdot)$ and $\text{cos}(\cdot)$  functions in \eqref{eq:SimpleForm} with their polar representations from \eqref{eq:QuaternionSinCos}, gives
\begin{equation}
\begin{aligned}
q_{\left[n\right]}=&\underbrace{\Big(\frac{A_{\left[n\right]}e^{\zeta''(\phi_{\zeta,\left[n\right]})}}{2\zeta''}+\frac{B_{\left[n\right]}e^{\zeta''(\phi_{\zeta',\left[n\right]})}}{2}\Big)e^{\zeta''(2 \pi f \Delta T n )}\zeta}_{q^{+}_{\left[n\right]}}
\\
&-\underbrace{\Big(\frac{A_{\left[n\right]}e^{-\zeta''(\phi_{\zeta,\left[n\right]})}}{2\zeta''}+\frac{B_{\left[n\right]}e^{-\zeta''(\phi_{\zeta',\left[n\right]})}}{2}\Big)e^{-\zeta''(2 \pi f \Delta T n )}\zeta}_{q^{-}_{\left[n\right]}}
\end{aligned}
\label{eq:ThreePhaseSignal}
\end{equation}
where $q^{+}_{\left[n\right]}= e^{\zeta''(2 \pi f \Delta T)}q^{+}_{\left[n-1\right]} \hspace{0.32cm} \text{and} \hspace{0.32cm} q^{-}_{\left[n\right]}= e^{-\zeta''(2 \pi f \Delta T)}q^{-}_{\left[n-1\right]}$ trace two counter-rotating circles in the imaginary subspace of $\mathbb{H}$. Note that ideally, $q^{-}_{[n]}=0$, referred to as an balanced system. These two counter rotating circles are shown in Fig.~\ref{Fig:Plot2}. From \eqref{eq:ThreePhaseSignal}, state of the system is represented via the vector 
\begin{equation}	\begin{bmatrix}\varphi_{\left[n+1\right]}&q^{+}_{\left[n+1\right]}&q^{-}_{\left[n+1\right]}\end{bmatrix}^{\T}=\begin{bmatrix}\varphi_{\left[n\right]}& \varphi_{\left[n\right]} q^{+}_{\left[n\right]}& \varphi^{*}_{\left[n\right]}q^{-}_{\left[n\right]}\end{bmatrix}^{\T}
\label{eq:QuaternionStateThreePhase}
\end{equation} 
where $\varphi_{\left[n\right]}=e^{\zeta'' 2 \pi f_{\left[n\right]} \Delta T }$ with $f_{\left[n\right]}$ denoting the instantaneous frequency of the system. The quaternion model has two key advantages over its complex-valued  duals. First, it allows to distinguish between different classes of  faults~\cite{PouriaPhD}. Second, it produces estimates of system frequency with very low bias. This is demonstrated in Fig.~\ref{Fig:BiasMSE}, where state of the system is tracked using the quaternion Kalman filter and the model in \eqref{eq:QuaternionStateThreePhase} with frequency of the system found from
\begin{equation}
\hat{f}_{\left[n\right]}=\frac{1}{2\pi\Delta T}\text{atan}\left(\frac{\|\Im(\hat{\varphi}_{n})\|}{\Re(\hat{\varphi}_{n})}\right)\cdot
\label{eq:QFE-with-noise}
\end{equation} 
Note that using a quaternion model not only results in a lower mean square error (MSE), but also bias of the obtained estimates are extremely small as well.   

\subsection{Linear Dynamic Programming and Initial Elements of Reinforcement Learning}

\label{Sec:LPRL}

Consider the optimisation problem formulated as:
\begin{align}
\text{minimise}:&\hspace{0.16cm}J=\mathbf{x}^{a\H}_{\left[N\right]}\mathbf{T}^{a}\mathbf{x}^{a}_{\left[N\right]}+\sum^{N-1}_{n=1}\left(\mathbf{x}^{a\H}_{\left[n\right]}\mathbf{Q}\mathbf{x}^{a}_{\left[n\right]}+\mathbf{u}^{a\H}_{\left[n\right]}\mathbf{R}\mathbf{u}^{a}_{\left[n\right]}\right)\label{eq:Control-Cost-Net}
\\
\text{so that:}&\hspace{0.16cm}\mathbf{u}_{\left[n\right]}=\mathbf{G}_{\left[n\right]}\mathbf{x}^{a}_{n}\hspace{0.26cm}\text{and}\hspace{0.26cm}\mathbf{x}^{a}_{\left[n+1\right]}=\mathbf{A}\mathbf{x}^{a}_{\left[n\right]}+\mathbf{B}\mathbf{u}^{a}_{\left[n\right]}\label{eq:Control-Subject}
\end{align}
where $\{\mathbf{T},\mathbf{R},\mathbf{Q}\}$ are Hermitian transpose symmetric and positive semi-definite, that is, $\forall\mathbf{z}\in\mathbb{H}^{M}\hspace{0.12cm}\&\hspace{0.12cm}\mathbf{M}\in\{\mathbf{T},\mathbf{R},\mathbf{Q}\}:\hspace{0.16cm}\mathbf{z}^{a\H}\mathbf{M}\mathbf{z}^{a}\in\mathbb{R}^{+}\cup\{0\}$. The aim is to find inputs, $\left\{\mathbf{u}_{\left[1\right]},\ldots,\mathbf{u}_{\left[N\right]}\right\}$, to dynamic system,  $\mathbf{x}^{a}_{\left[n+1\right]}=\mathbf{A}\mathbf{x}^{a}_{\left[n\right]}+\mathbf{B}\mathbf{u}^{a}_{\left[n\right]}$, so that cost function in \eqref{eq:Control-Cost-Net} is minimised. In this context, $\mathbf{G}_{\left[n\right]}$ is the feedback gain matrix, referred to as optimal policy in reinforcement learning problems~\cite{RL}. 

The solution requires calculation of the optimal feedback gain matrix at each time instant. To this end, the cost function in \eqref{eq:Control-Cost-Net} is sectioned into time specific cost functions given by
\begin{equation}
J_{\left[n\right]}=\left\{\begin{array}{ll}\mathbf{x}^{a\H}_{\left[N\right]}\mathbf{T}\mathbf{x}^{a}_{\left[N\right]}&\text{if }n=N,\\\mathbf{x}^{a\H}_{\left[n\right]}\mathbf{Q}\mathbf{x}^{a}_{\left[n\right]}+\mathbf{u}^{a\H}_{\left[n\right]}\mathbf{R}\mathbf{u}^{a}_{\left[n\right]}&\text{otherwise.}
\end{array}\right.
\label{eq:time-local-costs}
\end{equation}
Now, upon inserting \eqref{eq:Control-Subject} into \eqref{eq:time-local-costs} we have
\begin{equation}
J_{\left[n\right]}=\mathbf{x}^{a\H}_{\left[n\right]}\mathbf{Q}\mathbf{x}^{a}_{\left[n\right]}+\mathbf{u}^{a\H}_{\left[n\right]}\mathbf{R}\mathbf{u}^{a}_{\left[n\right]}=\mathbf{x}^{a\H}_{\left[n\right]}\boldsymbol{\Psi}_{\left[n\right]}\mathbf{x}^{a}_{\left[n\right]}
\label{eq:Time-Cost}
\end{equation}
with $\boldsymbol{\Psi}_{\left[n\right]}=\mathbf{Q}+\mathbf{G}^{\H}_{\left[n\right]}\mathbf{R}\mathbf{G}_{\left[n\right]}$. The expressions in \eqref{eq:time-local-costs} and \eqref{eq:Time-Cost} allow the time accumulative cost function to be defined as
\begin{equation}
\mathscr{J}_{\left[n\right]}=\sum^{N}_{m=n}J_{\left[m\right]}.
\label{eq:ac-cost}
\end{equation}
This indicates that there exists a positive definite Hermitian symmetric matrix $\mathbf{P}_{\left[n\right]}$, the existence of which is proven in SM-6, such that
\begin{equation}
\mathscr{J}_{\left[n\right]}=\mathbf{x}^{a\H}_{\left[n\right]}\mathbf{P}_{\left[n\right]}\mathbf{x}^{a}_{\left[n\right]}. 
\label{eq:HBJE}
\end{equation}
This extends the Hamilton-Jacobi-Bellman recursions (see~\cite{Stengel}) for solving quaternion-valued optimisation problems.  

Moving backwards in time with the implicit assumption that inputs for future time instances have been obtained, the task becomes that of calculating the input at the previous time instant resulting in
\begin{equation}
\mathscr{J}_{\left[n\right]}=\underset{\mathbf{u}_{\left[n\right]}}{\text{min}}\left\{\mathbf{x}^{a\H}_{\left[n\right]}\mathbf{Q}\mathbf{x}^{a}_{\left[n\right]}+\mathbf{u}^{\H}_{\left[n\right]}\mathbf{R}\mathbf{u}_{\left[n\right]}+\mathscr{J}_{\left[n\right]}\right\}.
\label{eq:quat-Bellman}
\end{equation}
The minimum of (\ref{eq:quat-Bellman}) occurs at
\begin{equation}	\mathbf{u}_{\left[n\right]}=-\left(\mathbf{R}+\mathbf{B}^{\H}\mathbf{P}_{\left[n+1\right]}\mathbf{B}\right)^{-1}\mathbf{B}^{\H}\mathbf{P}_{\left[n+1\right]}\mathbf{A}\mathbf{x}^{a}_{\left[n\right]}.
\label{eq:control-input-stack}
\end{equation}
Thus, from \eqref{eq:control-input-stack} and \eqref{eq:Control-Subject} it becomes clear that
\begin{equation}
\mathbf{G}_{\left[n\right]}=-\left(\mathbf{R}+\mathbf{B}^{\H}\mathbf{P}_{\left[n+1\right]}\mathbf{B}\right)^{-1}\mathbf{B}^{\H}\mathbf{P}_{\left[n+1\right]}\mathbf{A}.
\label{eq:OptimalG}
\end{equation}
An example for the use of quaternion linear dynamic programming for aircraft control is presented in SM-7. However, next we will delve into a more recent development, that is the use of quaternions in quantum computing. 

\begin{figure}[h!]
\includegraphics[width=0.9\linewidth,trim = 0cm 1.2cm 0cm 0cm]{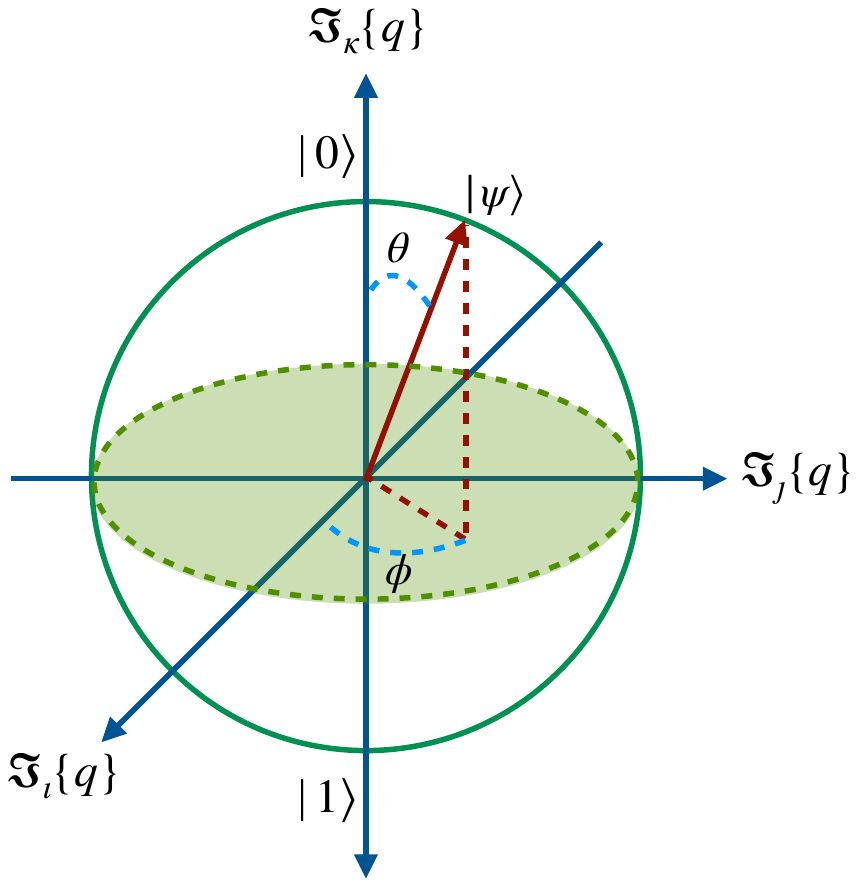}
\caption{A qubit representation with Bloch sphere, where $|\psi\rangle\leftrightarrow q=\imath\sin(\theta)\cos(\phi)+\jmath\sin(\theta)\sin(\phi)+ \kappa\cos(\theta)$.}
\label{Fig:BlochSphere}
\end{figure}

The fundamental unit of information in quantum computing is referred to as a quantum bit or qubit. In contrast to a classical bit that can only exist in a state of one or zero, a qubit can exist in both states simultaneously. This results from the principles of quantum superposition. Denoting state of zero (\textit{cf}. one) with $|0\rangle$ (\textit{cf}. $|1\rangle$), a qubit can be in any superposition that satisfies
\begin{equation} 
\forall\{\alpha,\beta\}\subset\mathbb{C}\wedge\|\alpha\|^{2}+\|\beta\|^{2}=1:\hspace{0.26cm}|\psi\rangle=\alpha|0\rangle+\beta|1\rangle
\end{equation}
where $|\psi\rangle$ represents state of the qubit while $\|\alpha\|^{2}$ and $\|\beta\|^{2}$ correspond to the probability of finding the qubit in state $|0\rangle$ or $|1\rangle$.\footnote{Note that qubit states arise from quantum characteristics of a particle, e.g. the up or down spin of an electron.}  Although use of two complex-valued variables $\alpha$ and $\beta$ might suggest four degrees of freedom for $|\psi\rangle$, the condition, $\|\alpha\|^{2}+\|\beta\|^{2}=1$, limits the degrees of freedom to three. Thus, as demonstrated in Fig.~\ref{Fig:BlochSphere}, the state of $|\psi\rangle$ can be described via a pure imaginary quaternion
\begin{equation}
|\psi\rangle\leftrightarrow q=\imath\sin(\theta)\cos(\phi)+\jmath\sin(\theta)\sin(\phi)+ \kappa\cos(\theta)
\end{equation} 
with angles $\theta$ and $\phi$ presented in Fig.~\ref{Fig:BlochSphere}. This representation is referred to as Bloch sphere. In general, an operation of a quantum computer corresponds to manipulations of the quantum state, or groups of quantum states. These operations are effectively modelled in the quaternion domain using the quaternion rotations as formulated in \eqref{eq:QuaternionRotation}, state-space methods, and $\mathbb{HR}$-calculus tools. For example, let us consider the quantum compiler problem. 

The input and output of quantum computers are unit pure imaginary quaternions, the setting derived in this article allows all quantum computing operations to be modelled effectively using the division algebra of quaternions, e.g. quaternion-valued involution and rotation. To this end, consider the general quantum computing operation represented by the mapping  
\begin{equation}
\mathbf{q}_{\text{out}}\leftarrow\mathrm{M}\left(\mathbf{q}_{\text{in}}\right)
\end{equation}
where $\mathrm{M}\left(\cdot\right)$ is a quaternion-valued function. On the most fundamental level, we would like to emulate this operation using a quantum computer the size and operations of which are restricted. Operation of this quantum computer is represented via $  \hat{\mathbf{q}}_{\text{out}}\leftarrow\hat{\mathrm{M}}\left(\mathbf{q}_{\text{in}},\mathbf{W}\right)$ with $\mathbf{W}$ denoting a parameter matrix determining the characteristics of $\hat{\mathrm{M}}\left(\cdot\right)$. This problem can be formulated as
\begin{align}
\min_{\mathbf{W}}&\left\{\|\mathrm{f}\left(\mathbf{q}_{\text{in}}\right)-\hat{\mathrm{f}}\left(\mathbf{q}_{\text{in}},\mathbf{W}\right)\|^{2}+\mathrm{g}\left(\mathbf{W}\right)\right\}\label{eq:QuantumLearn}
\\
\text{so that:}&\hspace{0.24cm}\mathbf{W}\in\mathcal{W}\label{eq:QuantumCondition}
\end{align}
where $\mathrm{g}\left(\cdot\right):\mathbb{H}^{M\times M}\rightarrow\mathbb{R}^{+}$ is a function representing the cost of implementing a certain structure and $\mathcal{W}$ is the set of admissible values for $\mathbf{W}$.  The goal is to find $\mathbf{W}$ so that the implementable mapping, $\hat{\mathrm{M}}\left(\cdot\right)$, resembles that of the desired mapping $\mathrm{M}\left(\cdot\right)$. The solution to this problem has been considered in the form of factorisation and search algorithms~\cite{QuantComplier}. However, the parameter matrix can be found  using the $\mathbb{HR}$-calculus, to indicate the steepest direction of reduction in cost. On a simpler level, the minimum of \eqref{eq:QuantumLearn} is calculable via gradient descent. This solution can then be projected to the closest $\mathbf{W}\in\mathcal{W}$ in order to meet the condition in \eqref{eq:QuantumCondition}.  

\begin{rem}
The most important application of the setting derived in this article for modelling qubit is that it allows for most concepts in linear/nonlinear dynamic programming to be used in the context of quantum computing. Note that these concepts are expandable to the quaternion domain using the $\mathbb{HR}$-calculus.  For example, recent publications in the field of quantum machine learning use the quaternion presentation of qubits, similar to the setting in this article~\cite{QML1,QML2,QuantComplier}. However, for these approaches to become applicable, a framework for backpropagation in the quaternion domain is required which is presented in SM-5.
\end{rem}

\section{Conclusion}
	
In this article, an overview of key concepts in the $\mathbb{HR}$-calculus that play an integral role in derivation of quaternion-valued information processing techniques have been presented. Moreover, these concepts have been used to formulate a number of adaptive processing techniques, such as the LMS and Kalman filters, in the quaternion domain and demonstrate their advantages. The main vantage point of this approach is that the derived techniques are inherently quaternion-valued, preserving the notion of phase and the division algebra of quaternions. Finally, note that in this article only the concepts of derivatives and gradients have been considered, while formulation of integrals in this framework remains an open issue.

\bibliographystyle{IEEEbib}
	
\bibliography{ref}
	
{
		
\footnotesize
		
\noindent\textbf{Danilo Mandic} (d.mandic@ic.ac.uk) received the Ph.D. degree in nonlinear adaptive signal processing in 1999 from Imperial College, London, where he is now a Professor. He specialises in Statistical Learning Theory, Machine Intelligence, and Statistical Signal Processing. Recently, he was invited as a distinguished lecturer by the IEEE Signal Processing Society to talk on a wide range of topics ranging from graph theory to innovation in education.
		
\noindent\textbf{Sayed Pouria Talebi} (s.talebi12@alumni.imperial.ac.uk) received his doctorate degree from Imperial College London, where is research focused on quaternion-valued signal processing. He has served as a researcher at a number of institutions including University of Cambridge. He is currently a lecturer at University of Roehampton. Some of his contributions to this area include formulation of the theory of quaternion-valued multi-agent systems, formulation of the basis of quaternion-valued control theory, and quaternion-valued nonlinear/non-Gaussian signal processing. In his work, he has demonstrated the usefulness of quaternions for modelling power distributions systems and passive target tracking. 
		
\noindent\textbf{Clive Cheong Took} (clive.cheongtook@rhul.ac.uk) received his doctorate from Cardiff University. He is a senior lecturer at Royal Holloway, University of London, U.K. He has been working on quaternion signal processing since 2008 and delivered a tutorial on this topic in the conference EUSIPCO 2011. Some of his contributions to this area include augmented statistics, adaptive learning, and novel matrix factorisations. Over the years, Clive has demonstrated the usefulness of quaternions to address various problems such as source separation, prediction or estimation, and classification for various biomedical applications. Recently, his research in quaternions has nurtured his interest in three-dimensional rotations for which quaternions are popular for. He is a Senior Member of IEEE.
		
\noindent\textbf{Yili Xia} (yili\textunderscore xia@seu.edu.cn)  received the Ph.D. degree in adaptive signal processing from Imperial College London in 2011. Since 2013, he has been an Associate Professor in signal processing with the School of Information Science and Engineering, Southeast University, where he is currently the Deputy Head of the Department of Information and Signal Processing Engineering. His research interests include complex and hyper-complex statistical analysis. He was a recipient of the Best Student Paper Award at the International Symposium on Neural Networks (ISNN) in 2010 (coauthor) and the Education Innovation Award at the IEEE International Conference on Acoustics, Speech, and Signal Processing (ICASSP) in 2019. He is an Associate Editor for the IEEE Transactions on Signal Processing.
		
\noindent\textbf{Dongpo Xu} (dongpoxu@gmail.com) received the Ph.D. degree in computational mathematics from the Dalian University of Technology, Dalian, China in 2009. He was a Lecturer with the College of Science, Harbin Engineering University and a Visiting Scholar with the Department of Electrical and Electronic Engineering, Imperial College London. He is currently a Professor with the School of Mathematics and Statistics, Northeast Normal University, Changchun, China. His current research interests include neural networks, machine learning, and signal processing. Dr. Xu has been an Associate Editor of the Frontiers in Artificial Intelligence and the Frontiers in Big Data.
		
\noindent\textbf{Min Xiang} (xiang\_min@buaa.edu.cn) received the Ph.D. degree in electrical engineering from Imperial College London, London, U.K., in 2018. He was a Research Associate with Imperial College London from 2018 to 2020. Since 2020, he has been an Associate Professor with Beihang University, Beijing, China. His research interests include statistical signal processing, specifically quaternion-valued signal processing.
		
\noindent\textbf{Pauline Bourigault} (p.bourigault22@imperial.ac.uk) is currently pursuing her doctoral degree in AI and Machine Learning enrolled in the UKRI Centre for Doctoral Training in AI for Healthcare at Imperial College London. Her current research focuses on 3D human pose estimation.
		
}
	

\newpage

\section*{Supplementary Material}

\subsection*{SM-1: Quaternion Rotations and Algebraic Lyapunov/Riccati Recursions}

\label{SM:LR}

Consider the evolution of a quaternion-valued sequence $\{q_{\left[n\right]}:n\in\mathbb{N}\}$ that encapsulates all three transformations (rotation, scaling, and shift) of a real-world system operating in the three-dimensional space, e.g.,  robot arm,  aircraft/vehicle, or even spin of an quantum particle. Based on the quaternion model of rotation as given in \eqref{eq:QuaternionRotation}, the evolutions of  $q_{\left[n\right]}$ can be formulated as 
\begin{equation}
q_{\left[n+1\right]}=a_{\left[n\right]}\mu_{\left[n\right]}q_{\left[n\right]}\mu^{-1}_{\left[n\right]}+q_{\text{shift}_{\left[n\right]}}=\sqrt{a_{\left[n\right]}}\mu_{\left[n\right]}q_{\left[n\right]}\sqrt{a_{\left[n\right]}}\mu^{-1}+q_{\text{shift}_{\left[n\right]}}
\label{eq:LyaponouvQuaternion}
\end{equation}
where $a_{\left[n\right]}\in\mathbb{R}$ and $q_{\text{shift}_{\left[n\right]}}$ represent the scaling and shift factors. Moreover, as an associative hypercomplex algebra, quaternions have an isomorphic matrix algebra~\cite{QMA}, so that 
\begin{equation}
\forall  q\in\mathbb{H}\leftrightarrow\mathbf{Q}=\begin{bmatrix}q_{r}&-q_{\imath}&-q_{\jmath}&q_{\kappa}\\q_{\imath}&q_{r}&-q_{\kappa}&-q_{\jmath}\\q_{\jmath}&q_{\kappa}&q_{r}&q_{\imath}\\-q_{\kappa}&q_{\jmath}&-q_{\imath}&q_{r}\end{bmatrix}\in\mathbb{R}^{4}\times\mathbb{R}^{4}
\label{eq:matrix-dual}
\end{equation}
where $\mathbf{Q}$ is the representation of $q$ in the isomorphic matrix algebra~\cite{QMA}. Thus, using the duality in \eqref{eq:matrix-dual} the expression in \eqref{eq:LyaponouvQuaternion} can be transformed into the real-valued Lyapunov style recursion 
\begin{equation}
\mathbf{Q}_{\left[n+1\right]}=\mathbf{M}_{\left[n\right]}\mathbf{Q}_{\left[n+1\right]}\mathbf{M}^{\H}_{\left[n\right]}+\mathbf{Q}_{\text{shift}_{\left[n\right]}}
\end{equation}
with $\mathbf{Q}_{\left[n\right]}$, $\mathbf{M}_{\left[n\right]}$, and $\mathbf{Q}_{\text{shift}_{\left[n\right]}}$ representing the real-valued matrix duals of $q_{\left[n\right]}$, $\sqrt{a}\mu$, and $q_{\text{shift}_{\left[n\right]}}$. In addition, consider the Riccati recursion in its information matrix formulation given as
\begin{equation}
\mathbf{Q}_{\left[n+1\right]}=\underbrace{\left(\underbrace{\left(\mathbf{M}_{\left[n\right]}\mathbf{Q}_{\left[n\right]}\mathbf{M}^{\H}_{\left[n\right]}+\mathbf{Q}_{\text{shift}_{\left[n\right]}}\right)^{-1}}_{\text{inner Lyapunov operator}}+\mathbf{H}^{\H}_{\left[n\right]}\mathbf{R}^{-1}_{\left[n\right]}\mathbf{H}_{\left[n\right]}\right)^{-1}}_{\text{outer Lyapunov operator}}
\label{eq:QuaternionRiccati}
\end{equation}
for general matrices $\mathbf{H}_{\left[n\right]}$ and $\mathbf{R}_{\left[n\right]}$. Then, it becomes clear that \eqref{eq:QuaternionRiccati} consists of an inner and an outer Lyapunov style operators that interact, which essentially models two interacting systems of the from in \eqref{eq:LyaponouvQuaternion}. These dualities highlight an important issue. The Lyapunov and Riccati recursions used to model the involutions of physical systems are essentially representing the rotation, scaling, and shift, that these systems experience from one time instant to the next, which are more effectively dealt with using quaternions. Finally, the example given for a scalar quaternion-valued random variable is extendable to higher dimensional systems. For more details on the use and behaviour of quaternion-valued Lyapunov and Riccati recursions, the reader is referred to~\cite{PouriaPhD,QuaternionControl}.

\subsection*{SM-2: Taylor Series Expansion and Linearisation}

\label{SM:Taylor}

Consider the first-order Taylor series expansion in its real-valued formulation given by
\begin{equation}
\mathrm{f}\left(\mathbf{x}+\Delta\mathbf{x}\right)=\mathrm{f}\left(\mathbf{x}\right)+\left(\Delta\mathbf{x}\right)^{\T}\left(\nabla_{\mathbf{x}}\mathrm{f}\left(\mathbf{x}\right)\right)+\mathcal{O}\left(\mathrm{f}\left(\mathbf{x}\right)\right)
\label{eq:RealTaylor}
\end{equation}
where $\mathbf{x}^{\T}=\left[\mathbf{x}^{\T}_{r},\mathbf{x}^{\T}_{\imath},\mathbf{x}^{\T}_{\jmath},\mathbf{x}^{\T}_{\kappa}\right]$,  $\Delta\mathbf{x}$ represents the change in $\mathbf{x}$, and $\mathcal{O}\left(\mathrm{f}\left(\mathbf{x}\right)\right)$ denotes higher-order terms. Now, mapping $\mathbf{x}$, $\Delta\mathbf{x}$, and $\nabla_{\mathbf{x}}\mathrm{f}\left(\mathbf{x}\right)$ on to the quaternion domain using \eqref{eq:Observation-Model} and \eqref{eq:GradientMapping} allows \eqref{eq:RealTaylor} to be formulated in the quaternion domain as~\cite{Cyrus}
\begin{equation}
\mathrm{f}\left(\mathbf{x}^{a}+\Delta\mathbf{x}^{a}\right)= \mathrm{f}\left(\mathbf{x}^{a}\right)+\left(\Delta\mathbf{x}^{a}\right)^{\H}\left(\nabla_{\mathbf{x}^{a*}}\mathrm{f}\right)+\mathcal{O}\left(\mathrm{f}\left(\mathbf{x}^{a}\right)\right).
\end{equation}
One can formulate higher-order expansion of the series in a similar manner~\cite{GenHR}. Finally, note that the Taylor series expansion is formulated to be linear with respect to $\mathbf{x}^{a}$, which is useful in the context of linearisation and adaptive information processing techniques~\cite{Cyrus,PouriaPhD}. It must be mentioned that alternative approaches exist~\cite{Clif}. Finally, a detailed derivation of Taylor's theorem in the quaternion domain can be found in~\cite{Clif,GenHR}.

\subsection*{SM-3: Real-Valued Components of \eqref{eq:TheThing}}

Given $\{\mathrm{f}\left(\cdot\right),\mathrm{g}\left(\cdot\right)\}:\mathbb{H}^{M}\rightarrow\mathbb{H}$ the real-valued components of $\mathrm{h}\left(\mathbf{q}\right)=\mathrm{f}\left(\mathbf{q}\right)\mathrm{g}\left(\mathbf{q}\right)$ are
\begin{align}
\Re\left\{\mathrm{h}\left(\mathbf{q}\right)\right\}=&\left(\Re\left\{\mathrm{f}\left(\mathbf{q}\right)\right\}\right)\left(\Re\left\{\mathrm{g}\left(\mathbf{q}\right)\right\}\right)+\left(\Im_{\imath}\left\{\mathrm{f}\left(\mathbf{q}\right)\right\}\right)\left(\Im_{\jmath}\left\{\mathrm{g}\left(\mathbf{q}\right)\right\}\right)
\\
&+\left(\Im_{\jmath}\left\{\mathrm{f}\left(\mathbf{q}\right)\right\}\right)\left(\Im_{\jmath}\left\{\mathrm{g}\left(\mathbf{q}\right)\right\}\right)+\left(\Im_{\kappa}\left\{\mathrm{f}\left(\mathbf{q}\right)\right\}\right)\left(\Im_{\kappa}\left\{\mathrm{g}\left(\mathbf{q}\right)\right\}\right)\nonumber
\\
\Im_{\imath}\left\{\mathrm{h}\left(\mathbf{q}\right)\right\}=&\left(\Re\left\{\mathrm{f}\left(\mathbf{q}\right)\right\}\right)\left(\Im_{\imath}\left\{\mathrm{g}\left(\mathbf{q}\right)\right\}\right)+\left(\Im_{\imath}\left\{\mathrm{f}\left(\mathbf{q}\right)\right\}\right)\left(\Re\left\{\mathrm{g}\left(\mathbf{q}\right)\right\}\right)
\\
&+\left(\Im_{\jmath}\left\{\mathrm{f}\left(\mathbf{q}\right)\right\}\right)\left(\Im_{\kappa}\left\{\mathrm{g}\left(\mathbf{q}\right)\right\}\right)+\left(\Im_{\kappa}\left\{\mathrm{f}\left(\mathbf{q}\right)\right\}\right)\left(\Im_{\jmath}\left\{\mathrm{g}\left(\mathbf{q}\right)\right\}\right)\nonumber
\\
\Im_{\jmath}\left\{\mathrm{h}\left(\mathbf{q}\right)\right\}=&\left(\Re\left\{\mathrm{f}\left(\mathbf{q}\right)\right\}\right)\left(\Im_{\jmath}\left\{\mathrm{g}\left(\mathbf{q}\right)\right\}\right)+\left(\Im_{\imath}\left\{\mathrm{f}\left(\mathbf{q}\right)\right\}\right)\left(\Im_{\kappa}\left\{\mathrm{g}\left(\mathbf{q}\right)\right\}\right)
\\
&+\left(\Im_{\jmath}\left\{\mathrm{f}\left(\mathbf{q}\right)\right\}\right)\left(\Re\left\{\mathrm{g}\left(\mathbf{q}\right)\right\}\right)+\left(\Im_{\kappa}\left\{\mathrm{f}\left(\mathbf{q}\right)\right\}\right)\left(\Im_{\imath}\left\{\mathrm{g}\left(\mathbf{q}\right)\right\}\right)\nonumber
\\
\Im_{\kappa}\left\{\mathrm{h}\left(\mathbf{q}\right)\right\}=&\left(\Re\left\{\mathrm{f}\left(\mathbf{q}\right)\right\}\right)\left(\Im_{\kappa}\left\{\mathrm{g}\left(\mathbf{q}\right)\right\}\right)+\left(\Im_{\imath}\left\{\mathrm{f}\left(\mathbf{q}\right)\right\}\right)\left(\Im_{\jmath}\left\{\mathrm{g}\left(\mathbf{q}\right)\right\}\right)
\\
&+\left(\Im_{\jmath}\left\{\mathrm{f}\left(\mathbf{q}\right)\right\}\right)\left(\Im_{\kappa}\left\{\mathrm{g}\left(\mathbf{q}\right)\right\}\right)+\left(\Im_{\kappa}\left\{\mathrm{f}\left(\mathbf{q}\right)\right\}\right)\left(\Re\left\{\mathrm{g}\left(\mathbf{q}\right)\right\}\right).\nonumber
\end{align}

\subsection*{SM-4: Quaternion-Valued Nonlinear Adaptive Filtering and Learning}

Consider the operations of a computational element (neuron or perceptron), formulated as
\begin{equation}
\hat{y}_{n}=\mathrm{h}\left(\mathbf{w}^{\T}_{\left[n\right]}\mathbf{z}^{a}_{\left[n\right]}\right)
\end{equation}
where $\mathbf{z}_{\left[n\right]}$ is a vector presenting the inputs, $\mathbf{w}_{\left[n\right]}$ is a vector of the corresponding weights, and $\mathrm{h}\left(\cdot\right):\mathbb{H}\rightarrow\mathbb{H}$ is a differentiable function in $\mathbb{H}$. Note that for mathematical presentation the bias element is simply considered as an element of $\mathbf{z}_{\left[n\right]}$. The aim is to find optimal weight vector, $\mathbf{w}^{\text{opt}}$, so that the output, $\hat{y}_{\left[n\right]}$, is a close fit to desired output, $y_{\left[n\right]}$, in a manner that minimises a selected metric function $\mathrm{d}\left(\cdot\right)$ so that $\left\{\mathbb{H},\mathrm{d}\left(\cdot\right)\right\}$ constitutes a convex metric space.

The weight vector is updated iteratively in order to reduce the cost
\begin{equation}
\mathrm{g}\left(\mathbf{w}_{\left[n\right]}\right)=\mathrm{d}\left(\hat{y}_{\left[n\right]}-y_{\left[n\right]}\right).
\end{equation}
Specifically, we would like to achieve
\begin{equation}
\mathrm{g}\left(\mathbf{w}_{\left[n+1\right]}\right)<\mathrm{g}\left(\mathbf{w}_{\left[n\right]}\right)\label{eq:cost-fucntion-fisrt}
\end{equation}
with $\mathbf{w}_{\left[n+1\right]}=\mathbf{w}_{\left[n\right]}+\Delta\mathbf{w}_{\left[n\right]}$ where $\Delta\mathbf{w}_{\left[n\right]}$ represents a change in the weight vector from time instant $n$ to time instant $n+1$. Taking the quaternion-valued Taylor series expansion of \eqref{eq:cost-fucntion-fisrt} yields
\begin{equation}
\mathrm{g}\left(\mathbf{w}_{\left[n\right]}\right)+\left(\Delta\mathbf{w}^{a}_{\left[n\right]}\right)^{\H}\nabla_{\mathbf{w}^{a*}_{\left[n\right]}}\mathrm{g}(\mathbf{w}_{\left[n\right]})+\mathcal{O}\left(\mathrm{g}\left(\mathbf{w}_{n}\right)\right)<\mathrm{g}\left(\mathbf{w}_{n}\right).
\label{eq:NFT}
\end{equation}
Neglecting the higher-order terms in \eqref{eq:NFT} gives $\left(\Delta\mathbf{w}^{a}_{\left[n\right]}\right)^{\H}\nabla_{\mathbf{w}^{a*}_{\left[n\right]}}\mathrm{g}(\mathbf{w}_{\left[n\right]})<0$. Therefore, considering $\Delta\mathbf{w}^{a}_{\left[n\right]}\propto\gamma\nabla_{\mathbf{w}^{a*}_{\left[n\right]}}\mathrm{g}(\mathbf{w}_{\left[n\right]})$ allows the formulation of a weight update procedure as
\begin{equation}
\mathbf{w}^{a}_{\left[n+1\right]}=\mathbf{w}^{a}_{\left[n\right]}-\gamma\nabla_{\mathbf{w}^{a*}_{\left[n\right]}}\mathrm{g}\left(\mathbf{w}_{\left[n\right]}\right)
\label{eq:first-order-update}
\end{equation}
where $\gamma\in\mathbb{R}^{{}^{+}}$ is an adaptation gain. 

\begin{rem}
Since $ \mathrm{d}\left(\cdot\right)$ is a metric function it is positive semi-definite. Thus, the selected adaptation gain should guarantee
\begin{equation}
\mathrm{g}\left(\mathbf{w}_{[n+1]}\right)=\mathrm{g}\left(\mathbf{w}_{\left[n\right]}\right)+\left(\Delta\mathbf{w}^{a}_{\left[n\right]}\right)^{\H}\nabla_{\mathbf{w}^{a*}_{\left[n\right]}}\mathrm{g}(\mathbf{w}_{\left[n\right]})+\mathcal{O}\left(\mathrm{g}\left(\mathbf{w}_{n}\right)\right)\geq0
\label{eq:FirstProofNonlinear}
\end{equation}
where we have used the first-order Taylor expansion of $\mathrm{g}\left(\mathbf{w}_{\left[n+1\right]}\right)$. Neglecting the higher-order terms of \eqref{eq:FirstProofNonlinear} and replacing $\left(\Delta\mathbf{w}^{a}_{\left[n\right]}\right)^{\H}$ with 
\begin{equation}
\left(\Delta\mathbf{w}^{a}_{\left[n\right]}\right)^{\H}=
\mathbf{w}^{a\H}_{\left[n+1\right]}-\mathbf{w}^{a\H}_{\left[n\right]}=-\gamma\left(\nabla_{\mathbf{w}^{a*}_{\left[n\right]}}\mathrm{g}\left(\mathbf{w}_{\left[n\right]}\right)\right)^{\H}
\end{equation}
obtained from the expression in  \eqref{eq:first-order-update}, yields
\begin{equation}
\mathrm{g}\left(\mathbf{w}_{[n+1]}\right)=\mathrm{g}\left(\mathbf{w}_{\left[n\right]}\right)-\gamma\left\|\nabla_{\mathbf{w}^{a*}_{\left[n\right]}}\mathrm{g}(\mathbf{w}_{\left[n\right]})\right\|^{2}_{2}\geq0.
\label{eq:condition}
\end{equation}
In order to guarantee the condition in \eqref{eq:condition}, it is required that
\begin{equation}
\gamma\leq\frac{\mathrm{g}\left(\mathbf{w}_{\left[n\right]}\right)}{\left\|\nabla_{\mathbf{w}^{a*}_{\left[n\right]}}\mathrm{g}(\mathbf{w}_{\left[n\right]})\right\|^{2}_{2}}\cdot
\end{equation}
For the special case of $\mathrm{d}\left(\cdot\right)=\|\cdot\|^{2}_{2}$ we have $\gamma\in\left(0,\frac{1}{8\|\mathbf{z}^{a}_{\left[n\right]}\|^{2}_{2}\|\nabla_{\mathbf{w}^{a*}_{\left[n\right]}}\mathrm{h}\left(\mathbf{w}^{T}\mathbf{z}^{a}_{\left[n\right]}\right)\|^{2}_{2}}\right)$.

\end{rem}

\subsection*{SM-5: Backpropagation in the Quaternion Domain}

Consider a neural network of $L$ layers with $N_{l}$ neurons in the $l$th layer. Output of the $n$th neuron on the $l$th layer can be formulated as
\begin{equation}
y_{l,n} = \sum_{m=1}^{N_{l}}w_{l,n,m}x_{l-1,m} + b_{l,n}
\end{equation}
where $w_{l,n,m}$ are quaternion-valued weights and $b_{l,n}$ is the bias term, while 
\begin{equation}
x_{l,n}= \mathrm{f}\left(y_{l,n}\right)
\end{equation}
is the output of neuron $n$ in layer $l$ with $\mathrm{f}\left(\cdot\right)$ representing a nonlinear quaternion-valued activation function. The aim is to find the gradient of the cost function
\begin{equation}
\mathrm{J}=\frac{1}{2}\sum_{m=1}^{N_L}\|d_{m} - x_{L,m}\|^{2} 
\label{eq:BPC}
\end{equation}
with respect to the quaternion-valued weights, where $d_{m}$ denotes a desired output. Given the cost function $\mathrm{J}$ in \eqref{eq:BPC}, the gradient is given by
\begin{equation}
\nabla_{\mathbf{W}^{*}_{L}}\mathrm{J} = -\frac{1}{2}\sum_{\forall\zeta\in\{1,\imath,\jmath,\kappa\}} \left(\nabla_{\mathbf{W}^{\zeta}_{L}}\mathrm{J}^{\H}\boldsymbol{\epsilon}\right)^{\zeta}
\end{equation}
with 
\begin{equation}
\mathbf{W}_{L}=\begin{bmatrix}w_{L,1,1}&\dots&w_{L,1,N_{L-1}}\\\vdots&\ddots&\vdots\\w_{L,N_{L},1}&\dots&w_{L,N_{L},N_{L-1}}\end{bmatrix}\hspace{0.26cm}\text{and}\hspace{0.26cm}
\boldsymbol{\epsilon}=\begin{bmatrix}d_{1}\\\dots\\d_{N_{L}}\end{bmatrix}-\underbrace{\begin{bmatrix}x_{L,1}\\\dots\\x_{L,N_{L}}\end{bmatrix}}_{\mathbf{x}_{L}}\cdot
\end{equation}
This gradient can be described in a layer-by-layer manner so that 
\begin{equation}
\boldsymbol{\delta}_{l}=\left\{\begin{array}{ll}\boldsymbol{\epsilon}&l=L\\\left(\mathbf{W}_{l+1}\mathbf{x}_{l+1}\right)^{\H}\boldsymbol{\delta}_{l+1} & l \neq L.\end{array}\right.
\end{equation}
The weight and bias update rules for the output layer then become
\begin{equation}
\mathbf{W}_{l} \leftarrow \mathbf{W}_{l} + \gamma\boldsymbol{\delta}_{l}\mathbf{x}^{\H}_{l-1}\hspace{0.26cm}\text{and}\hspace{0.26cm}\begin{bmatrix}b_{l,1}&\vdots&b_{l,N_{L}}\end{bmatrix}^{\T}\leftarrow\begin{bmatrix}b_{l,1}&\vdots&b_{l,N_{L}}\end{bmatrix}^{\T}+ \gamma\boldsymbol{\delta}_{l}
\end{equation}
while the weight and bias update rules for the hidden layers are given by
\begin{equation}
\mathbf{W}_{l} \leftarrow \mathbf{W}_{l} + \gamma\sum_{\forall\zeta\in \{1,\imath,\jmath,\kappa\}}\left(\boldsymbol{\delta}_{l} \right)^{\zeta} \mathbf{x}^{\H}_{l-1}\hspace{0.26cm}\text{and}\hspace{0.26cm}\begin{bmatrix}b_{l,1}\\\vdots\\ b_{l,N_{l}}\end{bmatrix}\leftarrow\begin{bmatrix}b_{l,1}\\\vdots\\ b_{l,N_{l}}\end{bmatrix}+\gamma\sum_{\forall\zeta\in\{1, i, j,k \}} \left(\boldsymbol{\delta}_{l} \right)^{\zeta}
\end{equation}
where $\gamma\in\mathbb{R}^{+}$ is the learning rate.

\begin{rem}
A survey of quaternion-valued neural networks, their applications, and their advantages is presented in~\cite{QNN,CompareQANN,Requested}. There are alternative methods for deriving training techniques for quaternion-valued neural networks, mainly based on the approaches in~\cite{Arena,Nitte} and \cite{ThesisQNN,BUCHHOLZ2008925} that consider quaternions as a special case of Clifford algebra. Although all techniques are shown to achieve similar performance, most recent contributions that deal with quaternion-valued signals use the $\mathbb{HR}$-calculus~\cite{CompareQANN,QBPP}.  
\end{rem}

\subsection*{SM-6: Quaternion-Valued Hamilton-Jacobi-Bellman Type Recursion}

The aim is to recover matrix sequence $\{\mathbf{P}_{\left[n\right]}:n=1,\ldots,N\}$ that stratifies \eqref{eq:ac-cost} and \eqref{eq:HBJE}. For the case of $n=N$, from (\ref{eq:ac-cost}) it follows that $\mathbf{P}_{\left[N\right]}=\mathbf{T}$. Now, we assume that a positive definite Hermitian symmetric matrix $\mathbf{P}_{\left[n\right]}$ exists, so that $\mathscr{J}_{\left[n\right]}=\mathbf{x}^{a\H}_{\left[n\right]}\mathbf{P}_{\left[n\right]}\mathbf{x}^{a}_{\left[n\right]}$. Then, for $\mathscr{J}_{\left[n-1\right]}$, we have
\begin{equation}
\begin{aligned}
\mathscr{J}_{\left[n-1\right]}=&\sum^{N}_{m=n-1}J_{\left[m\right]}=J_{\left[n-1\right]}+\underbrace{\sum^{N}_{m=n}J_{\left[m\right]}}_{\mathscr{J}_{\left[n\right]}}=\mathbf{x}^{a\H}_{\left[n-1\right]}\boldsymbol{\Psi}_{\left[n-1\right]}\mathbf{x}^{a}_{\left[n-1\right]}+\mathbf{x}^{a\H}_{\left[n\right]}\mathbf{P}^{a}_{\left[n\right]}\mathbf{x}^{a}_{\left[n\right]}.
\end{aligned}
\label{eq:cost-midway}
\end{equation}
Substituting $\mathbf{x}^{a}_{\left[n\right]}$ from \eqref{eq:Control-Subject}, into \eqref{eq:cost-midway}, we arrive at
\begin{equation}
\mathscr{J}_{\left[n-1\right]}=\mathbf{x}^{a\H}_{\left[n-1\right]}\mathbf{P}_{\left[n-1\right]}\mathbf{x}^{a}_{\left[n-1\right]}
\label{eq:cost-induction}
\end{equation}
with
\begin{equation}
\mathbf{P}_{\left[n-1\right]}=\boldsymbol{\Psi}_{\left[n-1\right]}+\left(\mathbf{A}^{\H}+\mathbf{G}^{\H}_{\left[n-1\right]}\mathbf{B}^{\H}\right)\mathbf{P}^{a}_{\left[n\right]}\left(\mathbf{A}+\mathbf{B}\mathbf{G}_{\left[n-1\right]}\right).
\label{eq:cost-too-much}
\end{equation}
Therefore, via induction, the expressions (\ref{eq:cost-midway}) and (\ref{eq:cost-induction}) show that there exists a positive definite matrix $\mathbf{P}_{\left[n-1\right]}$ so that $\mathscr{J}_{\left[n-1\right]}=\mathbf{x}^{a\H}_{\left[n-1\right]}\mathbf{P}_{\left[n-1\right]}\mathbf{x}^{a}_{\left[n-1\right]}$. 

\begin{rem}
Defining matrix $\mathbf{S}_{[n]}$ as the solution to
\begin{equation}
\mathbf{P}_{\left[n\right]}=\mathbf{A}^{\H}\mathbf{S}_{\left[n+1\right]}\mathbf{A}+\mathbf{Q}\end{equation}
and substituting \eqref{eq:cost-too-much} gives
\begin{equation}
\mathbf{S}_{\left[n\right]}=\left(\left(\mathbf{A}^{\H}\mathbf{S}_{\left[n+1\right]}\mathbf{A}+\mathbf{Q}\right)^{-1}+\mathbf{B}\mathbf{R}^{-1}\mathbf{B}^{\H}\right)^{-1}
\label{eq:cent-control}
\end{equation}
where we have used $\boldsymbol{\Psi}_{\left[n\right]}=\mathbf{Q}+\mathbf{G}^{\H}_{\left[n\right]}\mathbf{R}\mathbf{G}_{\left[n\right]}$ from Section~\ref{Sec:LPRL}. Note that \eqref{eq:cent-control} is a quaternion-valued algebraic Riccati recursion with termination condition $\mathbf{P}_{[N]}=\mathbf{T}$. For more information on Riccati recursions in the quaternion domain please see~SM-1 and~\cite{PouriaPhD,QuaternionControl}. Finally, the inputs can be formulated as
\begin{equation}
\mathbf{u}_{\left[n\right]}=-\mathbf{R}^{-1}\mathbf{B}^{\H}\mathbf{S}_{\left[n+1\right]}\mathbf{A}\mathbf{x}^{a}_{\left[n\right]}.
\label{eq:column-optima-vec}
\end{equation}
\end{rem} 

\begin{rem}
Note that the recursion that was the subject of this section is reminiscent of Hamilton-Jacobi-Bellman recursion derived in real-valued vector algebra~\cite{Stengel}, both in terms of their visual representation and in terms of its application.
\end{rem}

\subsection*{SM-7: Fly-by-Wire Flight Controller}

In their most general formulation, the equation of motion governing the rotation of solid bodies can be summarised as 
\begin{equation}
\frac{\partial \varphi}{\partial t}= \phi\hspace{0.4cm}\text{ and }\hspace{0.4cm}\frac{\partial \phi}{\partial t}=\nu
\label{eq:motion-equation}
\end{equation} 
where $\nu$ represents the input to the system, typically torque or a related variable concerning angular acceleration), while $\varphi$ represents the angular speed with $\phi$ indicating its rate of change. Transforming the equations in (\ref{eq:motion-equation}) into its discrete-time formulation yields
\begin{equation}
\begin{bmatrix}
\varphi_{\left[n+1\right]}\\ \phi_{\left[n+1\right]}\end{bmatrix}=\mathrm{f}\left(\begin{bmatrix}
\varphi_{\left[n+1\right]}\\ \phi_{\left[n+1\right]}\end{bmatrix},u_{\left[n\right]}\right)= \begin{bmatrix}1&\Delta T\\0&1\end{bmatrix}\begin{bmatrix}\varphi_{\left[n\right]}\\ \phi_{\left[n\right]}\end{bmatrix}+\begin{bmatrix}\Delta T^{2}/2 \\ \Delta T \end{bmatrix}u_{\left[n\right]}
\label{eq:state-track}
\end{equation}
where $\Delta T=\text{\unit[$0.04$]{s}}$ denotes the sampling interval. 

\begin{figure}[h!]
\centering
\includegraphics[width=0.9\linewidth,trim = 0cm 4.8cm 0cm 0cm]{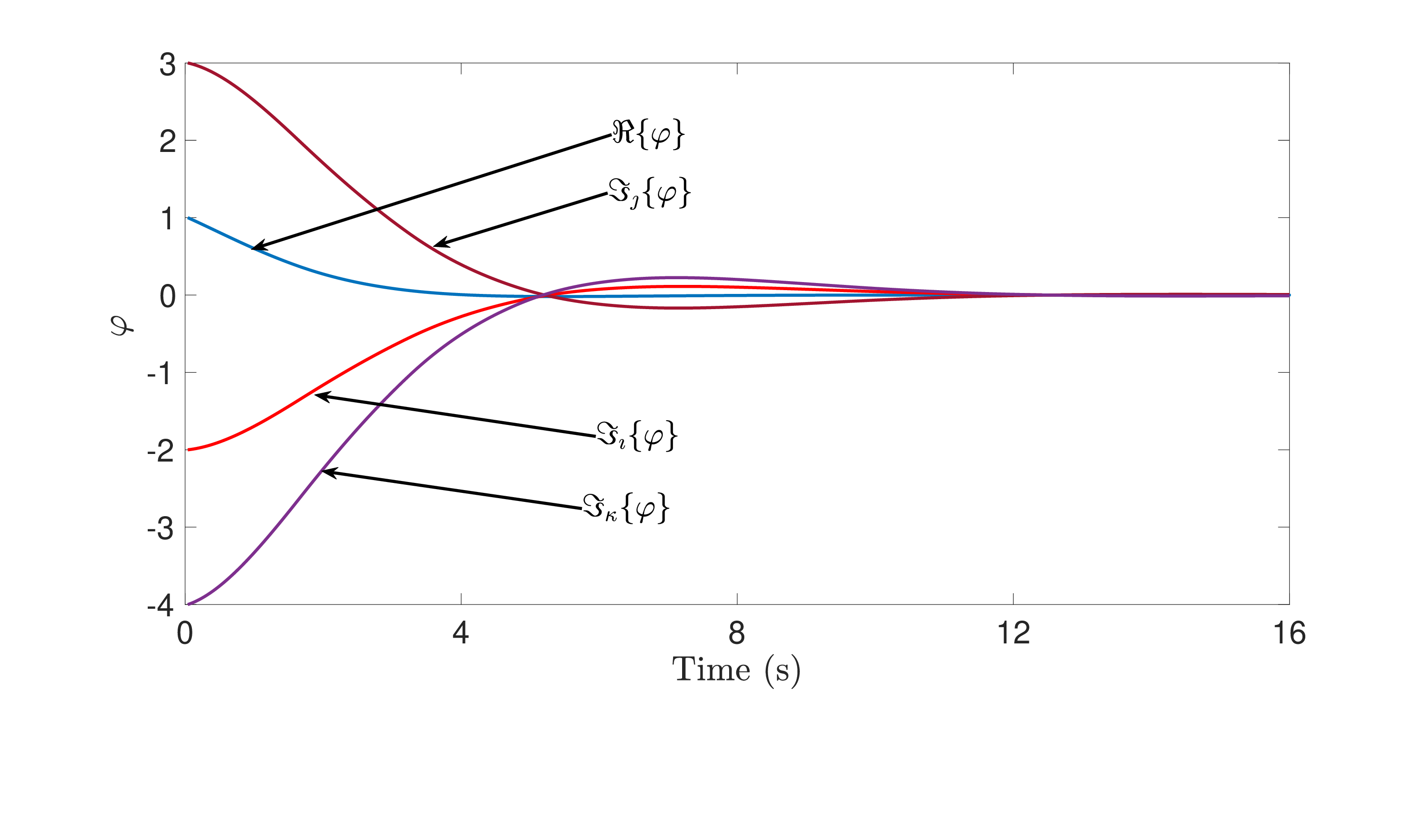}
\caption{Trajectory of $\varphi$ under the proposed controller scheme.}
\label{Fig:Control}
\end{figure}

In order to present an illustrative example, consider the input, $u_{n}$, to be linear feedback regulator found via the framework in Section~\ref{Sec:LPRL}, with
\begin{equation}
\mathbf{B}=\begin{bmatrix}\Delta T^{2}/2 \\ \Delta T \end{bmatrix}\text{and}\hspace{0.12cm}\hspace{0.12cm}\mathbf{A}= \begin{bmatrix}1&\Delta T\\0&1\end{bmatrix}
\end{equation}
while $\mathbf{Q}=\mathbf{I}$, $\mathbf{T}=50\times\mathbf{I}$, and $\mathbf{R}=10\times\mathbf{I}-1.875\times\left(\mathbf{1}\mathbf{1}^{\T}\right))$. Inputs were calculated for {\unit[$1.6$]{s}} long segments with the first half of the calculated sequences implemented. Then, input sequences were re-calculated using the new state-vector information. Fig.~\ref{Fig:Control} shows the smooth trajectory of $\varphi$. 

\begin{rem}
Note that quaternions have been an integral part of flight control systems~\cite{Survey}. Thus, this is not a novel application. However, quaternions have previously been decomposed into the real-part and a vector presenting the imaginary part. The advantage of the framework derived in this article, that is based on the use of the augmented quaternion vector, is twofold. First, the so-called augmented quaternion framework allows useful tools from algebra, calculus, and statistics to be recovered in the quaternion domain. Examples of this are the Taylor series expansion, and chain derivative rule as derived in SM-2 and Section~\ref{Sec:ChainRule}. In this article we derive a calculus for quaternion-valued functions, while~\cite{CliveSPM} uses the augmented quaternion vector to deal with quaternion-valued statistics. Second, the approach presented in this article removes the need for transformations, under which the elegance, physical interpretation, and division algebra of quaternions are lost or obscured. 
\end{rem}

\subsection*{SM-8: Statistical Gradient Descent}

A detailed account of quaternion statistics can be found in our sister publication~\cite{CliveSPM}~and~\cite{AQS,PouriaPhD}; however, for the purposes of this section we give brief introduction to the augmented quaternion characteristic function (AQCF). To this end, consider the real-valued random vector $\begin{bmatrix}\mathbf{q}^{\T}_{r}&\mathbf{q}^{\T}_{\imath}&\mathbf{q}^{\T}_{\jmath}&\mathbf{q}^{\T}_{\kappa}\end{bmatrix}^{\T}$ the characteristic function of which is defined as~\cite{ASB}
\begin{equation}
\mathbf{\Phi}_{\textbf{\textit{Q}}}(\mathbf{s}_{r},\mathbf{s}_{\imath},\mathbf{s}_{\jmath},\mathbf{s}_{\kappa})=\E{e^{\xi(\mathbf{s}^{\T}_{r}\mathbf{q}_{r}+\mathbf{s}^{\T}_{\imath}\mathbf{q}_{\imath}+\mathbf{s}^{\T}_{\jmath}\mathbf{q}_{\jmath}+\mathbf{s}^{\T}_{\kappa}\mathbf{q}_{\kappa})}}\cdot
\label{eq:CF-real}
\end{equation}
where $\xi$ is a unit pure imaginary number so that $\xi^{2}=-1$. Now, exploiting the relation in \eqref{eq:MappingHR}, we have 
\begin{equation}
\mathbf{q}^{a}=\underbrace{\begin{bmatrix}\mathbf{I}&\imath\mathbf{I}&\jmath\mathbf{I}&\kappa\mathbf{I}\\\mathbf{I}&\imath\mathbf{I}&-\jmath\mathbf{I}&-\kappa\mathbf{I}\\\mathbf{I}&-\imath\mathbf{I}&\jmath\mathbf{I}&-\kappa\mathbf{I}\\\mathbf{I}&-\imath\mathbf{I}&-\jmath\mathbf{I}&\kappa\mathbf{I}\end{bmatrix}}_{\mathbf{A}}\begin{bmatrix}\mathbf{q}_{r}\\\mathbf{q}_{\imath}\\\mathbf{q}_{\jmath}\\\mathbf{q}_{\kappa}\end{bmatrix}\hspace{0.18cm}\text{and}\hspace{0.18cm}
\mathbf{s}^{a}=\underbrace{\begin{bmatrix}\mathbf{I}&\imath\mathbf{I}&\jmath\mathbf{I}&\kappa\mathbf{I}\\\mathbf{I}&\imath\mathbf{I}&-\jmath\mathbf{I}&-\kappa\mathbf{I}\\\mathbf{I}&-\imath\mathbf{I}&\jmath\mathbf{I}&-\kappa\mathbf{I}\\\mathbf{I}&-\imath\mathbf{I}&-\jmath\mathbf{I}&\kappa\mathbf{I}\end{bmatrix}}_{\mathbf{A}}\begin{bmatrix}\mathbf{s}_{r}\\\mathbf{s}_{\imath}\\\mathbf{s}_{\jmath}\\\mathbf{s}_{\kappa}\end{bmatrix}
\label{eq:SQR}
\end{equation}
where substituting \eqref{eq:SQR} into \eqref{eq:CF-real}, gives formulation of the characteristic function in the quaternion domain as
\begin{equation}
\mathbf{\Phi}_{\textbf{\textit{Q}}^{a}}(\mathbf{s}^{a})=\E{e^{\xi\Re\{\mathbf{s}^{\H}\mathbf{q}\}}}=\E{e^{(\frac{\xi}{4}\mathbf{s}^{a\H}\mathbf{q}^{a})}}.
\label{eq:CF}
\end{equation}
Note that as \eqref{eq:CF} uses the quaternion random vector in its augmented form; thus, it is referred to as the augmented quaternion characteristic function (AQCF)~\cite{PouriaPhD}. In the reminder of this section we use the $\mathbb{HR}$-calculus and AQCF to establish the basis of statistical gradient descent algorithm in the quaternion domain.

\begin{rem}
All statistical moments, can be extracted from the derivatives of the AQCF. This is shown in SM-9. These moments are in turn useful in the derivation of statistical adaptive processing techniques. However, this requires the robust $\mathbb{HR}$-calculus framework. 
\end{rem}

\begin{rem}
Augmenting a quaternion-valued variable with its involution around the $\imath$, $\jmath$, and $\kappa$ axes, as formulated in \eqref{eq:MappingHR}, forms the foundations for augmented quaternion statistics and the $\mathbb{HR}$-calculus, as it provides a comprehensive method for four dimensional quaternion-valued variables to be considered from four rotational perspectives. 
\end{rem}

Let us formulate the problem in Section~\ref{Sec:Kalman} in a more general manner. The problem is that of tracking the state of an evolving system, whether a physical system (aircraft) or state of an adaptive processing machinery (weights of a quaternion-valued neural network). The state of this system is represented via the state vector sequence $\{\mathbf{x}_{\left[n\right]}: n\in\mathbb{N}\}$ so that
\begin{equation}
\mathbf{x}_{\left[n+1\right]}=\mathrm{f}\left(\mathbf{x}_{\left[n\right]},\mathbf{v}_{\left[n\right]},\mathbf{u}_{\left[n\right]}\right)
\label{eq:New-State-Model}
\end{equation}
where $\mathrm{f}(\cdot)$, $\mathbf{u}_{\left[n\right]}$, and $\mathbf{v}_{\left[n\right]}$, are the evolution function, a known control input, and a noise vector at time instant $n$. Informative observations regarding the state vector are assumed available through observations
\begin{equation}
\mathbf{y}_{\left[n\right]}=\mathrm{h}\left(\mathbf{x}_{\left[n\right]},\mathbf{w}_{\left[n\right]}\right)
\label{eq:New-Observation-Model}
\end{equation}
where $\mathrm{h}(\cdot)$ denotes the observation function, while  $\mathbf{y}_{\left[n\right]}$ and $\mathbf{w}_{\left[n\right]}$ are the observation and observation noise at time instant $n$. The aim is to find the estimate of $\mathbf{x}_{\left[n\right]}$ given the available information set $\mathbfcal{I}_{\left[n\right]}=\{\mathbf{u}_{\left[1:n\right]},\mathbf{y}_{\left[1:n\right]}\}$, formulated as $\hat{\mathbf{x}}_{\left[n\right]}=\E{\mathbf{x}_{\left[n\right]}|\mathbfcal{I}_{\left[n\right]}}$. 

From \eqref{eq:CF} and using \eqref{eq:DiffQuaternion}, we have 
\begin{equation}
\hat{\mathbf{x}}^{a}_{\left[n\right]}=\frac{4}{\xi}\frac{\partial\boldsymbol{\Phi}_{\textbf{\textit{x}}^{a}_{\left[n\right]}|\mathbfcal{I}_{\left[n\right]}}(\mathbf{s}^{a})}{\partial\mathbf{s}^{a*}}\mathlarger{\mathlarger{\mathlarger{|}}}_{\mathbf{s=0}}=\E{\mathbf{x}^{a}_{\left[n\right]}|\mathbfcal{I}_{\left[n\right]}}\cdot
\end{equation}
In the same manner, the \textit{a~priori} estimate is defined as
\begin{equation}
\boldsymbol{\psi}^{a}_{\left[n+1\right]}=\frac{4}{\xi}\frac{\partial\boldsymbol{\Phi}_{\textbf{\textit{x}}^{a}_{\left[n+1\right]}|\mathbfcal{I}_{\left[n\right]}}(\mathbf{s}^{a})}{\partial\mathbf{s}^{a*}}\mathlarger{\mathlarger{\mathlarger{|}}}_{\mathbf{s=0}}=\E{\mathbf{x}^{a}_{\left[n+1\right]}|\mathbfcal{I}_{\left[n\right]}}\cdot
\label{eq:Pre-Est-Def}
\end{equation}
where using \eqref{eq:State-Model} yields 
\begin{equation}
\boldsymbol{\Phi}_{\textbf{\textit{x}}^{a}_{\left[n+1\right]}|\mathbfcal{I}_{\left[n\right]}}(\mathbf{s}^{a})=\E{e^{\frac{\xi}{4}\mathbf{s}^{a\H}\mathbf{x}^{a}_{\left[n+1\right]}}|\mathbfcal{I}_{\left[n\right]}}=\E{e^{\frac{\xi}{4}\mathbf{s}^{a\H}\mathrm{f}^{a}(\mathbf{x}_{\left[n\right]},\mathbf{v}_{\left[n\right]},\mathbf{u}_{\left[n\right]})}|\mathbf{y}_{\left[1:n\right]}}\cdot
\label{eq:Pre-Est-Char}
\end{equation}
A substitution of \eqref{eq:Pre-Est-Def} into \eqref{eq:Pre-Est-Char} gives
\begin{equation}
\boldsymbol{\psi}^{a}_{\left[n+1\right]}=\E{\mathrm{f}^{a}(\mathbf{x}_{\left[n\right]},\mathbf{v}_{\left[n\right]},\mathbf{u}_{\left[n\right]})|\mathbf{y}_{\left[1:n\right]}}\approx\E{\mathrm{f}^{a}(\mathbf{x}_{\left[n\right]},\mathbf{u}_{\left[n\right]})|\mathbf{y}_{\left[1:n\right]}}
\label{eq:A-Postriori}
\end{equation}
allowing for an approximation of $\boldsymbol{\psi}^{a}_{\left[n+1\right]}$ to be formulated as
\begin{equation}
\boldsymbol{\psi}^{a}_{\left[n+1\right]}\approx\E{\mathbf{A}_{\left[n\right]}\hat{\mathbf{x}}^{a}_{\left[n\right]}+\mathcal{O}_{\mathrm{f}(\mathbf{x}_{\left[n\right]},\mathbf{u}_{\left[n\right]})}|\mathbf{y}_{\left[1:n\right]}}\approx\mathbf{A}_{\left[n\right]}\hat{\mathbf{x}}^{a}_{\left[n\right]}+\E{\mathcal{O}_{\mathrm{f}(\mathbf{x}_{\left[n\right]},\mathbf{u}_{\left[n\right]})}|\mathbf{y}_{\left[1:n\right]}}
\label{eq:A-Post-Approx}
\end{equation}
where $\mathbf{A}_{\left[n\right]}$ represents the linearisation of $\mathrm{f}(\cdot)$ at $\{\hat{\mathbf{x}}^{a}_{\left[n\right]},\mathbf{u}^{a}_{\left[n\right]}\}$ obtainable through the gradient operator \eqref{eq:GradientMapping}. In addition, we have 
\begin{equation}
\hat{\mathbf{x}}^{a}_{\left[n+1\right]}=\frac{4}{\xi}\frac{\partial\boldsymbol{\Phi}_{\mathbf{x}^{a}_{\left[n+1\right]}|\mathbfcal{I}_{\left[n+1\right]}}(\mathbf{s})}{\partial\mathbf{s}^{a*}}
\end{equation}
with $
\boldsymbol{\Phi}_{\mathbf{x}^{a}_{\left[n+1\right]}|\mathbfcal{I}_{\left[n+1\right]}}(\mathbf{s})=\E{e^{\frac{\xi}{4}\mathbf{s}^{a\H}\mathrm{f}^{a}(\mathbf{x}_{\left[n\right]},\mathbf{u}_{\left[n\right]})}e^{\frac{\xi}{4}\mathbf{s}^{a\H}\mathcal{O}_{(\mathrm{f}(\mathbf{x}_{\left[n\right]},\mathbf{v}_{\left[n\right]},\mathbf{u}_{\left[n\right]}))}}|\mathbf{y}_{\left[1:n+1\right]}}$.

Assuming that dependency of $\mathcal{O}_{(\mathrm{f}(\mathbf{x}_{\left[n\right]},\mathbf{v}_{\left[n\right]},\mathbf{u}_{\left[n\right]}))}$ on $\mathrm{f}(\mathbf{x}_{\left[n\right]},\mathbf{u}_{\left[n\right]})$ is negligible, \eqref{eq:A-Post-Approx} can be reformulated to give
\begin{equation}
\boldsymbol{\Phi}_{\mathbf{x}^{a}_{\left[n+1\right]}|\mathbfcal{I}_{\left[n+1\right]}}(\mathbf{s}^{a})=\boldsymbol{\Phi}_{\mathbf{x}^{a}_{\left[n+1\right]}|\mathbfcal{I}_{\left[n\right]}}(\mathbf{s}^{a})\E{e^{\frac{\xi}{4}\mathbf{s}^{a\H}\mathcal{O}_{(\mathrm{f}(\mathbf{x}_{\left[n\right]},\mathbf{v}_{\left[n\right]},\mathbf{u}_{\left[n\right]})}}|\mathbf{y}_{\left[n+1\right]}}
\end{equation}
where $\boldsymbol{\Phi}_{\mathbf{x}^{a}_{\left[n+1\right]}|\mathbfcal{I}_{\left[n+1\right]}}(\mathbf{s}^{a})$ consists of two parts, one representing the \textit{a~priori}, $\boldsymbol{\Phi}_{\mathbf{x}^{a}_{\left[n+1\right]}|\mathbfcal{I}_{\left[n\right]}}(\mathbf{s}^{a})$, and the other, representing the \textit{a~posteriori}, $\E{e^{j\mathbf{s}^{\T}\mathcal{O}_{(\mathrm{f}_{\left[n\right]}(\mathbf{x}_{\left[n\right]},\mathbf{v}_{\left[n\right]},\mathbf{u}_{\left[n\right]}))}}|\mathbf{y}_{\left[n+1\right]}}$, correction. Moreover, from \eqref{eq:A-Postriori}, we have
\begin{equation}
\hat{\mathbf{x}}^{a}_{\left[n+1\right]}=\boldsymbol{\psi}^{a}_{\left[n+1\right]}+\E{\mathcal{O}_{(\mathrm{f}_{\left[n\right]}(\mathbf{x}_{\left[n\right]},\mathbf{v}_{\left[n\right]},\mathbf{u}_{\left[n\right]}))}|\mathbf{y}_{\left[n+1\right]}}\cdot
\label{eq:Estimate-Update}
\end{equation}
Although the estimate, $\hat{\mathbf{x}}^{a}_{\left[n+1\right]}$, in \eqref{eq:Estimate-Update} can be approximated using particle based inference techniques, the $\mathbb{HR}$-calculus can provide a more computationally friendly and mathematically tractable solution. From \eqref{eq:New-State-Model} and \eqref{eq:New-Observation-Model}, the following error terms are defined
\begin{equation}
\boldsymbol{\epsilon}_{\left[n\right]}=\mathbf{x}_{\left[n\right]}-\hat{\mathbf{x}}_{\left[n\right]}\hspace{0.24cm}\text{and}\hspace{0.24cm}\tilde{\boldsymbol{\epsilon}}_{\left[n\right]}=\mathbf{y}_{\left[n\right]}-\hat{\mathbf{y}}_{\left[n\right]}
\label{eq:Error-Obs}
\end{equation}
where $\hat{\mathbf{y}}_{\left[n+1\right]}=\mathrm{h}(\boldsymbol{\psi}_{\left[n+1\right]})$ is the projection of $\boldsymbol{\psi}_{\left[n+1\right]}$ onto the observation space. Now, let $\mathrm{d}(\cdot)$ be a quaternion-valued metric function. Then, $\{\boldsymbol{\epsilon}_{\left[n\right]}:n\in\mathbb{N}\}$ being an exponentially stable sequence implies that $\{\tilde{\boldsymbol{\epsilon}}_{\left[n\right]}:n\in\mathbb{N}\}$ should also be exponentially stable.\footnote{Note that; i) converse of the statement might not hold true, and ii) the statement implicitly implies that $\mathrm{h}(\cdot)$ and $\mathrm{f}(\cdot)$ are bounded.} In this setting, $\nabla_{\boldsymbol{\psi}^{a*}_{\left[n+1\right]}}\mathrm{d}(\tilde{\boldsymbol{\epsilon}}_{\left[n+1\right]})$ indicates a direction of ascent for $\mathrm{d}(\tilde{\boldsymbol{\epsilon}}_{\left[n+1\right]})$, and hence, an adaptive filtering structure is formulated as
\begin{equation}
\hat{\mathbf{x}}^{a}_{\left[n+1\right]}\approx\boldsymbol{\psi}^{a}_{\left[n+1\right]}-\mathbf{G}_{\left[n\right]}\E{\nabla_{\boldsymbol{\psi}^{a*}_{\left[n+1\right]}}\mathrm{d}(\tilde{\boldsymbol{\epsilon}}_{\left[n+1\right]})}\approx\boldsymbol{\psi}^{a}_{\left[n+1\right]}-\mathbf{G}_{\left[n\right]}\nabla_{\boldsymbol{\psi}^{a*}_{\left[n+1\right]}}\mathrm{d}(\tilde{\boldsymbol{\epsilon}}_{\left[n+1\right]})
\label{eq:Adaptive-Filter}
\end{equation}
where $\{\mathbf{G}_{\left[n\right]}:n\in\mathbb{N}\}$ is a gain sequence while $\E{\nabla_{\boldsymbol{\psi}^{a*}_{\left[n+1\right]}}\mathrm{d}(\tilde{\boldsymbol{\epsilon}}_{\left[n+1\right]})}$ is replaced with its instantaneous value, $\nabla_{\boldsymbol{\psi}^{a*}_{\left[n+1\right]}}\mathrm{d}(\tilde{\boldsymbol{\epsilon}}_{\left[n+1\right]})$, as in most cases these expectations are not available. 

Most adaptive filtering techniques can be formalised within the framework derived in this section. For example, let the evolution, observation, and metric functions be defined as
\begin{equation}
\begin{aligned}
\mathrm{f}\left(\mathbf{x}_{\left[n\right]},\mathbf{v}_{\left[n\right]},\mathbf{u}_{\left[n\right]}\right)=&\mathbf{A}\mathbf{x}^{a}_{\left[n\right]}+\mathbf{v}^{a}_{\left[n\right]}
\\
\mathrm{h}\left(\mathbf{x}_{\left[n\right]},\mathbf{w}_{\left[n\right]}\right)=&\mathbf{H}\mathbf{x}^{a}_{\left[n\right]}+\mathbf{w}^{a}_{\left[n\right]}
\\
\mathrm{d}\left(\tilde{\boldsymbol{\epsilon}}_{\left[n\right]}\right)=&4\left\|\tilde{\boldsymbol{\epsilon}}_{\left[n\right]}\right\|^{2}=\tr{\tilde{\boldsymbol{\epsilon}}^{a}_{\left[n\right]}\tilde{\boldsymbol{\epsilon}}^{a\H}_{\left[n\right]}}.
\end{aligned}
\label{eq:SystemInformation}
\end{equation}
In this setting, from the framework of the $\mathbb{HR}$-calculus and \eqref{eq:Adaptive-Filter}, we have
\begin{equation}
\hat{\mathbf{x}}^{a}_{\left[n+1\right]}=\underbrace{\mathbf{A}\hat{\mathbf{x}}^{a}_{\left[n\right]}}_{\boldsymbol{\psi}^{a}_{\left[n+1\right]}}+\mathbf{G}_{\left[n\right]}\left(\mathbf{y}_{\left[n\right]}-\underbrace{\mathbf{H}\hat{\mathbf{x}}^{a}_{\left[n\right]}}_{\hat{\mathbf{y}}_{\left[n\right]}}\right)
\label{eq:StateEvo}
\end{equation}
which is the setting for the quaternion-valued Kalman filter in Section~\ref{Sec:Kalman}. In order to further demonstrate the role of the statistical gradient descent technique, we demonstrate its role in sensor fusion next.

Let us start with a set of sensors $\{\mathsf{S}_{l}:l=1,\ldots,M\}$ deployed to provide information regarding a system. For example, we can be interested in the state of a wide-area network, such as a power distribution network with voltage sensors in different locations, or sensor might be providing different information, such as a plane with sensors reading pitch, yaw, role, air speed, and position of control surfaces on the wings. Each sensor, $\mathsf{S}_{l}$, has estimate 
\begin{equation}\hat{\mathbf{x}}_{l,\left[n\right]}=\E{\mathbf{x}_{\left[n\right]}|\mathbf{y}_{l,\left[1:n\right]}}
\end{equation} 
with $\mathbf{y}_{l,\left[1:n\right]}$ representing the observations of sensor $l$ up to time instant $n$. The aim is to find $\hat{\mathbf{x}}_{\left[n\right]}=\E{\mathbf{x}_{\left[n\right]}|\hat{\mathbf{x}}_{1:M,\left[1:n\right]}}$. This is \textit{on~par} with the problem considered in \eqref{eq:New-State-Model} and \eqref{eq:New-Observation-Model}, so that; i) $\{\mathrm{f}\left(\cdot\right),\mathrm{h}\left(\cdot\right)\}$ become identity functions, ii) $\boldsymbol{\psi}_{\left[n\right]}$ becomes a vector incorporating any \textit{a~priori} information, and iii) $\begin{bmatrix}\hat{\mathbf{x}}^{a\T}_{1,\left[n\right]}&\ldots&\hat{\mathbf{x}}^{a\T}_{m,\left[n\right]}\end{bmatrix}^{\T}$ becomes the available observation at time instant $n$, i.e., $\mathbf{y}_{\left[n\right]}$. Using the the same machinery as in \eqref{eq:Pre-Est-Def}-\eqref{eq:Adaptive-Filter} and in the same format as  \eqref{eq:Adaptive-Filter} it follows that 
\begin{equation}
\hat{\mathbf{x}}^{a}_{\left[n+1\right]}=\boldsymbol{\psi}^{a}_{\left[n+1\right]}-\mathbf{G}_{\left[n\right]}\nabla_{\psi^{a}_{\left[n+1\right]}}\mathrm{d}\left(\psi^{a}_{\left[n+1\right]}-\mathbf{y}^{a}_{\left[n\right]}\right).
\label{eq:FirstSensor}
\end{equation}
Furthermore, it is generally assumed that; i) sensor estimates are independent with Gaussian distributions, ii) $\mathrm{d}\left(\cdot\right)=\|\cdot\|^{2}$. Thus, through the framework presented in Section~\ref{Sec:Kalman}, \eqref{eq:FirstSensor} simplifies into
\begin{equation}
\hat{\mathbf{x}}^{a}_{\left[n+1\right]}=\boldsymbol{\psi}^{a}_{\left[n+1\right]}-\mathbf{G}_{\left[n\right]}\left(\sum^{M}_{l=1}\boldsymbol{\Sigma}^{-1}_{\left(\mathbf{x}^{a}_{\left[n\right]}-\hat{\mathbf{x}}^{a}_{l,\left[n\right]}\right)}\hat{\mathbf{x}}^{a}_{\left[n\right]}\right)
\label{eq:SecondSensor}
\end{equation}
where $\boldsymbol{\Sigma}_{\left(\mathbf{x}^{a}_{\left[n\right]}-\hat{\mathbf{x}}^{a}_{l,\left[n\right]}\right)}=\E{\left(\mathbf{x}^{a}_{\left[n\right]}-\hat{\mathbf{x}}^{a}_{l,\left[n\right]}\right)\left(\mathbf{x}^{a}_{\left[n\right]}-\hat{\mathbf{x}}^{a}_{l,\left[n\right]}\right)^{\H}}$ and $\mathbf{G}_{\left[n\right]}$ is selected to minimise $	\boldsymbol{\Sigma}_{\left(\mathbf{x}^{a}_{\left[n\right]}-\hat{\mathbf{x}}^{a}_{\left[n\right]}\right)}$. Although the required augmented covariance matrix sequence, $\{	\boldsymbol{\Sigma}_{\left(\mathbf{x}^{a}_{\left[n\right]}-\hat{\mathbf{x}}^{a}_{l,\left[n\right]}\right)}:l=1,\ldots,M\}$, might be mathematically tractable (see~\cite{QuaternionControl}), it is generally assumed that $\boldsymbol{\psi}^{a}_{\left[n+1\right]}\rightarrow0$  and $\forall l:\boldsymbol{\Sigma}_{\left(\mathbf{x}^{a}_{\left[n\right]}-\hat{\mathbf{x}}^{a}_{l,\left[n\right]}\right)}\rightarrow w_{l}\mathbf{I}$, further simplifying \eqref{eq:SecondSensor} into 
\begin{equation}
\mathbf{x}^{a}_{\left[n+1\right]}=\sum^{M}_{n=1}w_{l}\mathbf{x}^{a}_{l,\left[n\right]}
\label{eq:Fusion}
\end{equation}
where we have assumed $\sum^{M}_{l=1}\mathbf{w}_{l}=1$ to enforce\footnote{Note that since the combination weight $w_{l}$ is taking the role of a positive definite covariance, and therefore, $w_{l}\in\mathbb{R}^{+}$.} $\mathbf{G}_{\left[n\right]}\rightarrow\mathbf{I}$.
Application of sensor fusion in distributed estimation and federated learning is elaborated upon next.

\noindent\textbf{1-Distributed Estimation}

\begin{figure}[h!]
\centering
\includegraphics[width=0.9\linewidth, trim = 0cm 2.4cm 0cm 0cm]{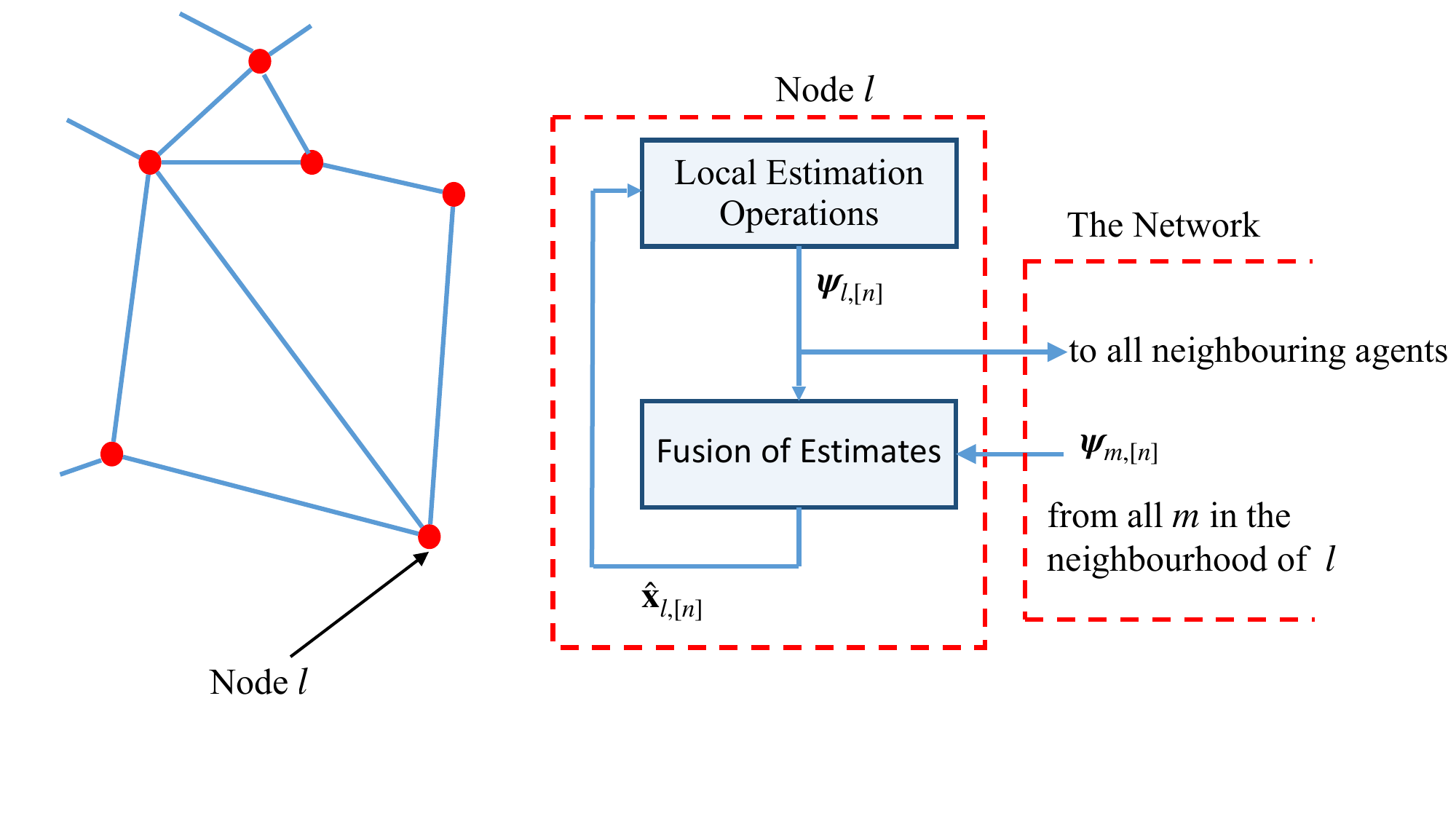}
\caption{Distributed learning scenario with a typical network shown on the left hand side  and operations of an agent shown on the right hand side. The agent formulates and shares intermediate estimates $\boldsymbol{\psi}_{l,\left[n\right]}$, while its fusion process produces estimates $\hat{\mathbf{x}}_{l,\left[n\right]}$.}
\label{Fig:DistributedEstimation}
\end{figure}

In distributed estimation settings, multiple agents cooperate over an ad-hoc\footnote{Form a network with ad-hoc structure, a decentralised type of network where agents or nodes connect directly to each other without the need for a centralised coordinator or pre-established network infrastructure. These networks are self-configuring, meaning that nodes can join or leave the network dynamically without relying on fixed infrastructure or administrative intervention.} communication network  in order to collaboratively estimate/track a state vector. In this setting, each agent implements a two stage learning process. First, based on \eqref{eq:Adaptive-Filter}, a local estimate is formed using only the observations of an agent. Second, the agents fuse their local estimates with those of their neighbours using \eqref{eq:Fusion}. This allows agents to learn from observations of other agents in the network, without the need for these observations to be shared. This scenario is shown in Fig.~\ref{Fig:DistributedEstimation}. 

\noindent\textbf{2-Federated Learning}

\begin{figure}[h!]
\centering
\includegraphics[width=0.9\linewidth,trim= 0cm 1.4cm 0cm 0cm]{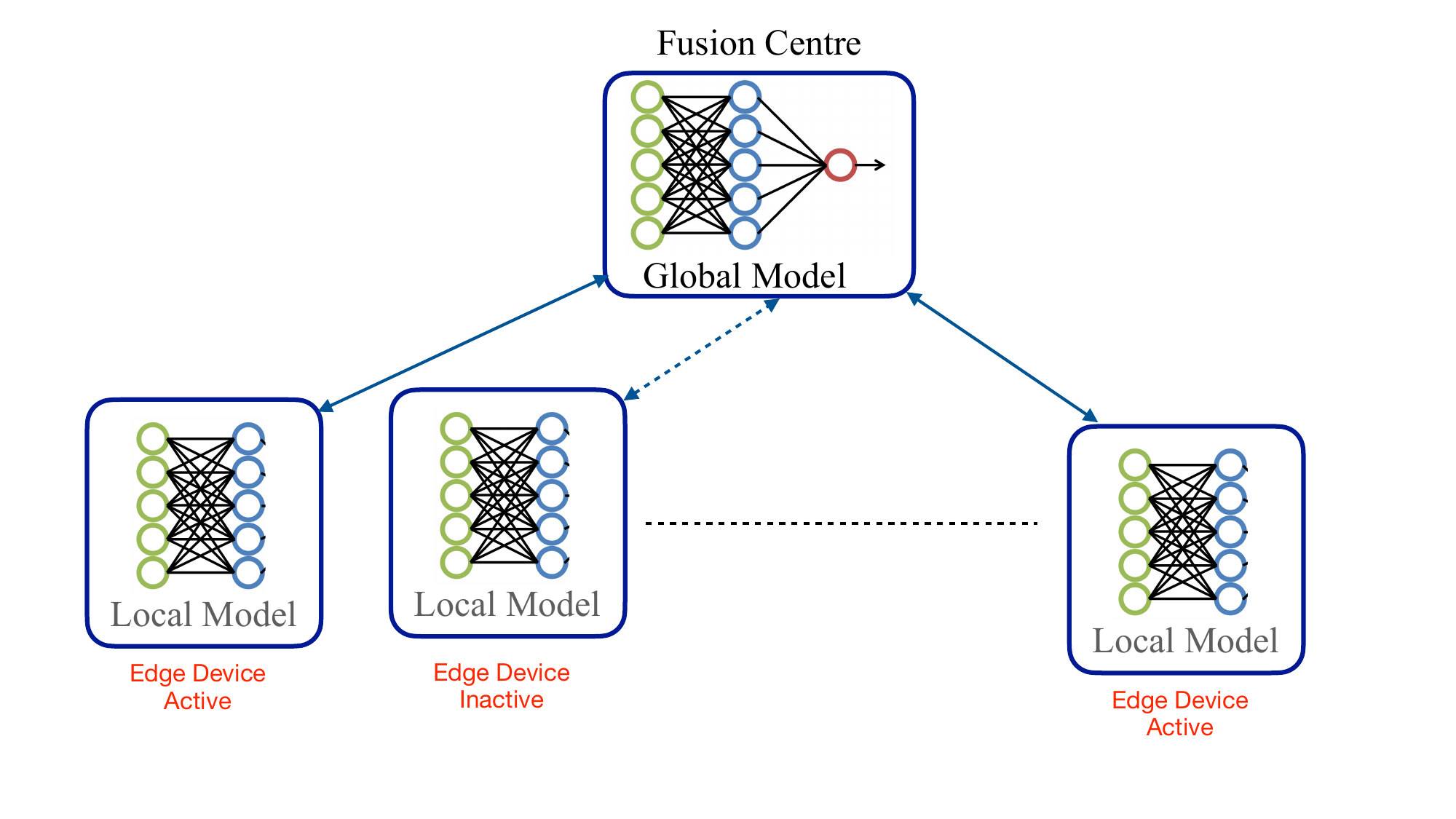}
\caption{The federated learning scenario is shown with a mix of active and inactive edge devices.}
\label{Fig:FedLearn}
\end{figure}

One of the most popular multi-agent scenarios for learning from data sets that are distributed among a multitude of agents is federated learning. On its most fundamental level, at each time instant, agents (referred to commonly as edge devices) that have new data sets available, update their estimates using \eqref{eq:Adaptive-Filter}, and then, send these updated estimates to a fusion centre which fuses these estimates with its existing information using \eqref{eq:SecondSensor}. Then, the fusion centre pushes these unified estimates of the learnt parameters back to the agents.  This scenario is shown in Fig.~\ref{Fig:FedLearn}.

In order to demonstrate application of sensor fusion in the quaternion domain, let us now consider a classical bearings-only-tracking problem. In this class of problems, the goal is to track the trajectory of a manoeuvring target using sensors that can only provide bearing measurements. This problem is often encountered in passive radar, sonar, or infrared tracking applications. As the distance to the target is not known the location of the target can only be resolved from combination of multiple sensor measurements and use of triangulation. However, the natural presentation of phase in quaternions allows for a much more elegant and effective approach to be implemented.

The target location at time instant $n$ is presented via the pure imaginary quaternion variable, $q_{\left[n\right]}$, in the manner given in Section~\ref{Sec:QuaternionRotation}. Then, dynamics of the target can be formulated as
\begin{equation}
\begin{aligned}
q_{\left[n+1\right]}=&q_{\left[n\right]}+v_{\left[n\right]}\Delta T + u_{\left[n\right]}\frac{1}{2}\left(\Delta T\right)^{2}
\\
v_{\left[n+1\right]}=&v_{\left[n\right]}+u_{\left[n\right]}\Delta T
\end{aligned}
\end{equation}
where $v_{\left[n\right]}$ and $u_{\left[n\right]}$ represent the speed and acceleration (normally modelled as zero-mean Gaussian noise) of the target, with $\Delta T$ representing the sampling interval. Moreover, sensor $l$ will have observations
\begin{equation}
y_{l,\left[n\right]}=\frac{q_{\left[n\right]}-L_{\mathsf{S}_{l}}}{\|q_{\left[n\right]}-L_{\mathsf{S}_{l}}\|}+w_{l,\left[n\right]}		
\end{equation}
where $w_{l,\left[n\right]}$ and $L_{\mathsf{S}_{l}}$ represent observation noise and location of sensor $l$. 

\begin{figure}[h!]
\centering
\includegraphics[scale=0.2]{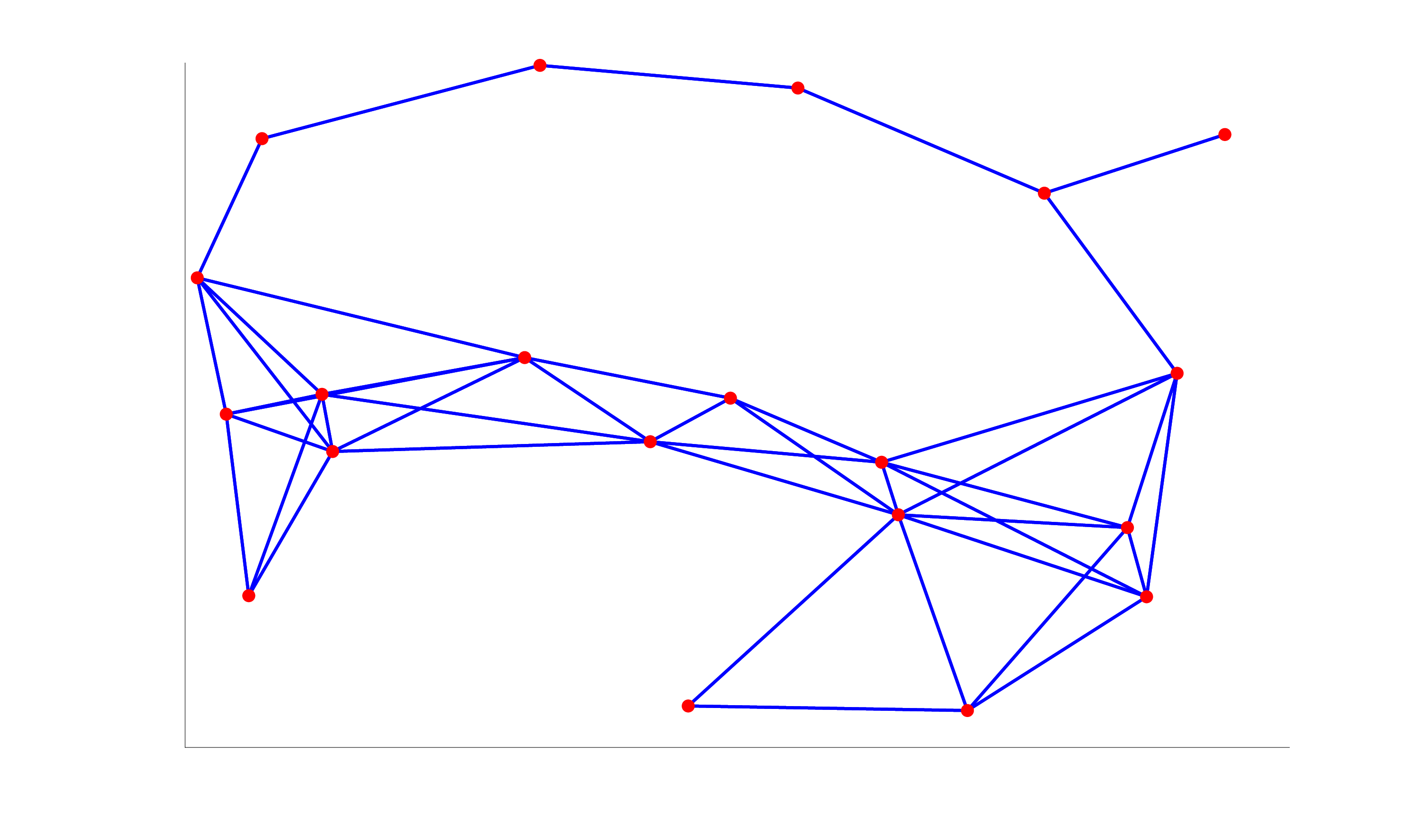}
\caption{Network of $20$ nodes and 43 connections used in simulations. Note that the figure only reflect the communication connection between the agents.}
\label{Fig:Network}
\end{figure}

\begin{figure}[h!]
\centering
\includegraphics[width=0.9\linewidth, trim = 0cm 0cm 0cm 0cm]{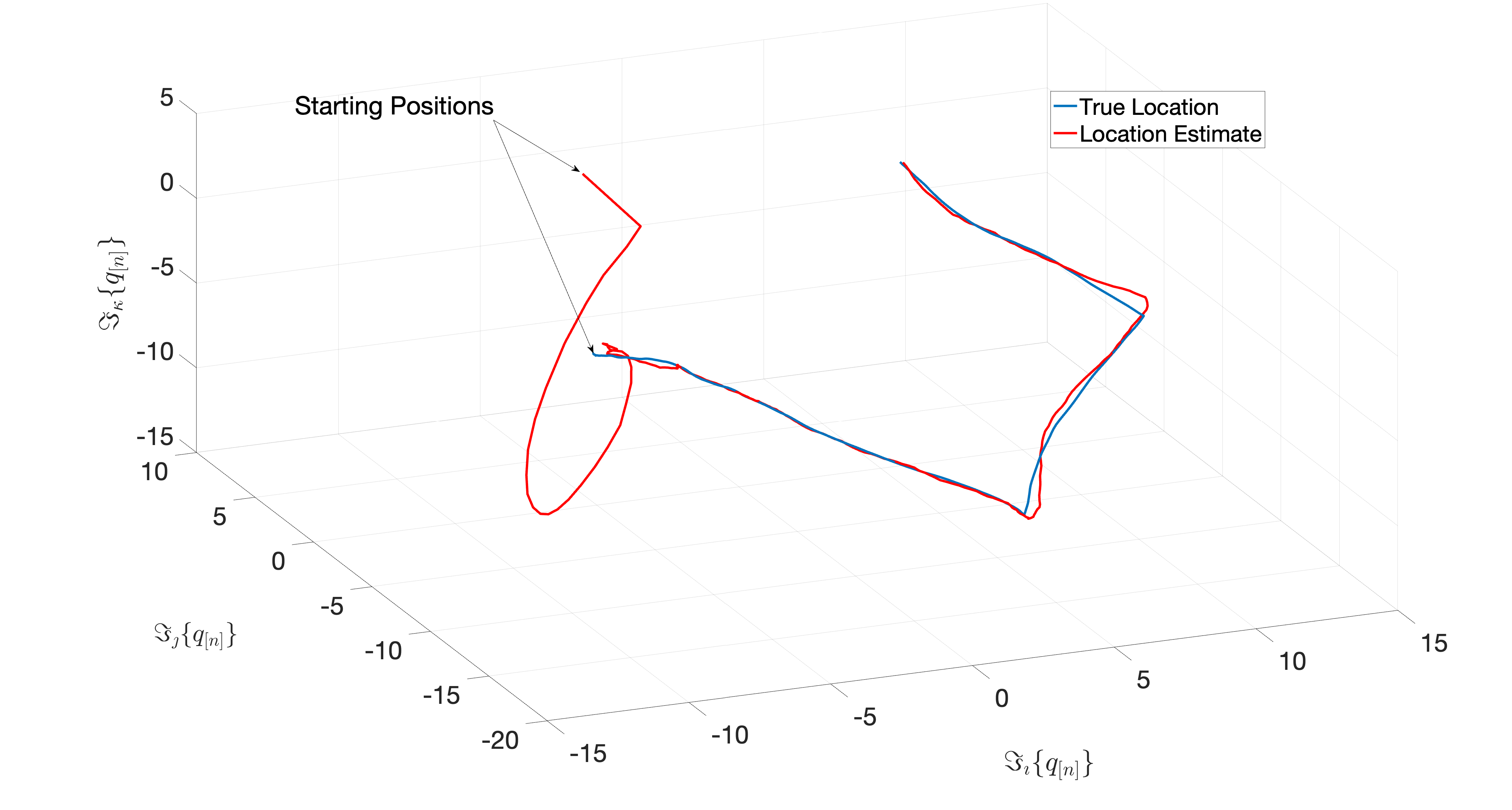}
\caption{Tracking solution of one agent in the bearings-only-tracking example.}
\label{Fig:Track}
\end{figure}

As an illustrative example, the network shown in Fig.~\ref{Fig:Network} with its nodes distributed uniformly (equally spaced) inside a $24\times24\times24$ cube was used to track a target moving inside the cube through bearings-only measurements. The sampling interval was set to $\Delta T = 0.04$~s. The acceleration was considered as a zero-mean Gaussian noise sequence with independent components and variance of $10$, while observation noise was considered to be zero-mean Gaussian with covariance of $10^{-4}$ and independent components.  Agents used framework of the stochastic gradient descent to formulate local estimates of the target which were then fused with those of the neighbouring agents via framework derived for sensor fusion. As a worst case scenario the tracking solution of the agent with only one connection is shown in Fig.~\ref{Fig:Track}.  Note the accurate tracking solution achieved via the division algebra of quaternions. Although a similar approach can be implemented in vector algebras, the lack of division operation means normalisation and sensor fusion operations require more complex computations, which often magnify estimation errors and compromise accuracy. 

\subsection*{SM-9: Relation Between the AQCF and Statistical Moments}

Consider the AQCF in the formulation given in (\ref{eq:CF}). Substituting the exponential with its power series representation yields
\begin{equation}
\begin{aligned}
\boldsymbol{\Phi}_{\textbf{\textit{Q}}^{a}}(\mathbf{s}^{a})=\E{e^{\left(\frac{\xi}{4}\mathbf{s}^{a\H}\mathbf{q}^{a}\right)}}=&\E{\sum^{\infty}_{m=0}\frac{\xi^{m}\left(\mathbf{s}^{a\H}\mathbf{q}^{a}\right)^{m}}{(4^{m})m!}}
\\
=&1+\E{\xi\frac{\mathbf{s}^{a\H}\mathbf{q}^{a}}{4}}-\E{\frac{\left(\mathbf{s}^{a\H}\mathbf{q}^{a}\right)^{2}}{32}}\cdots
\end{aligned}
\label{eq:CF-series}
\end{equation}
Using the $\mathbb{HR}$-calculus and derivatives in \eqref{eq:HR-derivatives}, from the expression in \eqref{eq:CF-series}, we have
\begin{equation}
\begin{aligned}
\frac{\partial\boldsymbol{\Phi}_{\textbf{\textit{Q}}^{a}}(\mathbf{s}^{a})}{\partial \mathbf{s}^{*\zeta}}=&\frac{\partial\boldsymbol{\Phi}_{\textbf{\textit{Q}}^{a}}(\mathbf{s}^{a})}{\partial\mathbf{s}^{a\H}\mathbf{q}^{a}}\frac{\partial \mathbf{s}^{a\H}\mathbf{q}^{a}}{\partial\mathbf{s}^{*\zeta}}=\E{\sum^{\infty}_{m=0}\frac{\xi^{m+1}\left(\mathbf{s}^{a\H}\mathbf{q}^{a}\right)^{m}}{(4^{m+1})m!}\mathbf{q}^{\zeta}}
\\
=&\E{\xi\frac{\mathbf{q}^{\zeta}}{4}}-\E{\frac{\mathbf{s}^{a\H}\mathbf{q}^{a}}{16}\mathbf{q}^{\zeta}}-\E{\xi\frac{\left(\mathbf{s}^{a\H}\mathbf{q}^{a}\right)^{2}}{128}\mathbf{q}^{\zeta}}\cdots
\end{aligned}
\label{eq:fAQCFd}
\end{equation}
with $\zeta\in\{1,\imath,\jmath,\kappa\}$. Evaluating \eqref{eq:fAQCFd} at $\mathbf{s}^{a}=0$ gives
\begin{equation}
\left.\frac{\partial\boldsymbol{\Phi}_{\textbf{\textit{Q}}^{a}}(\mathbf{s}^{a})}{\partial \mathbf{s}^{*\zeta}}\right|_{\mathbf{s}^{a}=0}=\E{\xi\frac{\mathbf{q}^{\zeta}}{4}}
\end{equation}
resulting in
\begin{equation}
\E{\mathbf{q}} = \left(\frac{4}{\xi}\left.\frac{\partial\boldsymbol{\Phi}_{\textbf{\textit{Q}}^{a}}(\mathbf{s}^{a})}{\partial \mathbf{s}^{*\zeta}}\right|_{\mathbf{s}^{a}=0}\right)^{\zeta}.
\end{equation}

For clarification, let us consider the example for $\zeta=1$, where we have
\begin{equation}
\begin{aligned}
 \frac{\partial\boldsymbol{\Phi}_{\textbf{\textit{Q}}^{a}}(\mathbf{s}^{a})}{\partial \mathbf{s}^{*}} =&\frac{\partial}{\partial\mathbf{s}^{*}}\E{\sum^{\infty}_{m=0}\frac{\xi^{m}\left(\mathbf{s}^{a\H}\mathbf{q}^{a}\right)^{m}}{(4^{m})m!}}=\E{\sum^{\infty}_{m=0}\frac{1}{(4^{m})m!}\frac{\partial\xi^{m}\left(\mathbf{s}^{a\H}\mathbf{q}^{a}\right)^{m}}{\partial\mathbf{s}^{*}}}
 \\
 =&\E{\sum^{\infty}_{m=0}\frac{\xi^{m}}{(4^{m})m!}\frac{\partial\left(\mathbf{s}^{a\H}\mathbf{q}^{a}\right)^{m}}{\partial\mathbf{s}^{(\xi^{m})^{-1}*}}}\cdot\hspace{0.12cm}\text{(using the rule in \eqref{eq:Important})}
\end{aligned}
\label{eq:FirstStepAQCF}
\end{equation}
Now, we focus on the main component of \eqref{eq:FirstStepAQCF}, that is
\begin{equation}
\frac{\partial\left(\mathbf{s}^{a\H}\mathbf{q}^{a}\right)^{m}}{\partial\mathbf{s}^{(\xi^{m})^{-1}*}}=m\left(\mathbf{s}^{a\H}\mathbf{q}^{a}\right)^{m-1}\frac{\partial\mathbf{s}^{a\H}\mathbf{q}^{a}}{\partial\mathbf{s}^{(\xi^{m})^{-1}*}}
\label{eq:AQCFFFF}
\end{equation}
where we have used the fact that $\mathbf{s}^{a\H}\mathbf{q}^{a}=4\Re\{\mathbf{s}^{\H}\mathbf{q}\}\in\mathbb{R}$ (see the derivation in \eqref{eq:CF-real}-\eqref{eq:CF}) and the rule in \eqref{eq:RealChain}. Once again considering that $\mathbf{s}^{a\H}\mathbf{q}^{a}=4\Re\{\mathbf{s}^{\H}\mathbf{q}\}\in\mathbb{R}$ and using the rule in \eqref{eq:RotationRule}, we have 
\begin{equation}
\frac{\partial\mathbf{s}^{a\H}\mathbf{q}^{a}}{\partial\mathbf{s}^{(\xi^{m})^{-1}*}}=\frac{\partial\mathbf{s}^{a\H}\mathbf{q}^{a}}{\partial\mathbf{s}^{*}}=\mathbf{q}.
\label{eq:LastStep}
\end{equation} 
Finally, replacing \eqref{eq:LastStep} into \eqref{eq:AQCFFFF} and \eqref{eq:FirstStepAQCF} yields the expression given in \eqref{eq:fAQCFd} for $\zeta=1$. The proof for general cases of $\zeta$ follows similarly and is detailed in~\cite{PouriaPhD}. 

\begin{rem}
In a similar fashion to the first-order derivatives, the second-order derivatives of $\mathbf{\Phi}_{\textbf{\textit{Q}}^{a}}\left(\mathbf{s}^{a}\right)$ can be expressed as
\begin{equation}
\frac{\partial^{2} \mathbf{\Phi}_{\textbf{\textit{Q}}^{a}}(\mathbf{s}^{a})}{\partial \mathbf{s}^{*\zeta_{1}}\partial\mathbf{s}^{*\zeta_{2}} }=\frac{\partial}{\partial\mathbf{s}^{*\zeta_{1}}}\left(\frac{\partial \mathbf{\Phi}_{\textbf{\textit{Q}}^{a}}(\mathbf{s}^{a})}{\partial\mathbf{s}^{*\zeta_{2}}}\right)=\frac{\partial}{\partial \mathbf{s}^{a\H}\mathbf{q}^{a}}\left(\frac{\partial \mathbf{\Phi}_{\textbf{\textit{Q}}^{a}}(\mathbf{s}^{a})}{\partial\mathbf{s}^{a\H}\mathbf{q}^{a}}\frac{\partial \mathbf{s}^{a\H}\mathbf{q}^{a}}{\partial\mathbf{s}^{*\zeta_{2}}}\right)\frac{\partial \mathbf{s}^{a\H}\mathbf{q}^{a}}{\partial \mathbf{s}^{*\zeta_{1}}}
\label{eq:sAQCFd}
\end{equation}
where $\zeta_{1},\zeta_{2} \in \{1,\imath,\jmath,\kappa\}$. Upon replacing (\ref{eq:fAQCFd}) into (\ref{eq:sAQCFd}) we have
\begin{equation}
\frac{\partial^{2} \mathbf{\Phi}_{\textbf{\textit{Q}}^{a}}(\mathbf{s}^{a})}{\partial \mathbf{s}^{*\zeta_{1}}\partial \mathbf{s}^{*\zeta_{2}} }=\E{\sum^{\infty}_{m=0}\frac{-\xi^{m}\left(\mathbf{s}^{a\H}\mathbf{q}^{a}\right)^{m}}{(4^{m+2})m!}\mathbf{q}^{\zeta_{2}}\left(\mathbf{q}^{\zeta_{1}}\right)^{\T}}
\end{equation}
that evaluated at $\mathbf{s}^{a}=0$ yields
\begin{equation}
\left.\frac{\partial^{2} \mathbf{\Phi}_{\textbf{\textit{Q}}^{a}}(\mathbf{s}^{a})}{\partial\mathbf{s}^{*\zeta_{1}}\partial\mathbf{s}^{*\zeta_{2}} }\right|_{\mathbf{s}^{a}=0}=\frac{-1}{16}\E{\mathbf{q}^{\zeta_{2}}\left(\mathbf{q}^{\zeta_{1}}\right)^{\T}}
\end{equation}
giving the cross-correlation between $\mathbf{q}^{\zeta_{1}}$ and $\mathbf{q}^{\zeta_{2}}$ multiplied by the factor $(\xi/4)^{2}$. Note that all higher-order statistical moments of random process $\textbf{\textit{Q}}$ are obtainable in an analogous manner. For exact derivation the reader is referred to~\cite{PouriaPhD}.
\end{rem}

\end{document}